\documentclass[conference]{IEEEtran}
\IEEEoverridecommandlockouts
\usepackage{cite}
\usepackage{amsmath,amssymb,amsfonts}
\usepackage{algorithmic}
\usepackage{graphicx}
\usepackage{textcomp}
\usepackage{xcolor}
\usepackage{multirow}
\usepackage{booktabs}
\usepackage{hyperref}
\usepackage{colortbl}
\usepackage{dsfont}

\def\BibTeX{{\rm B\kern-.05em{\sc i\kern-.025em b}\kern-.08em
    T\kern-.1667em\lower.7ex\hbox{E}\kern-.125emX}}

\makeatletter
\newcommand{\linebreakand}{%
  \end{@IEEEauthorhalign}
  \hfill\mbox{}\par
  \mbox{}\hfill\begin{@IEEEauthorhalign}
}
\makeatother

\begin{document}

\title{PiCo: Enhancing Text-Image Alignment \\with Improved Noise Selection and \\Precise Mask Control in Diffusion Models}

% \author{\IEEEauthorblockN{1\textsuperscript{st} Chang Xie}
% \IEEEauthorblockA{\textit{College of Computer Science and Technology} \\
% \textit{Nanjing University of Aeronautics and Astronautics} \\
% Nanjing, China\\
% xiechang@nuaa.edu.cn}
% \and
% \IEEEauthorblockN{1\textsuperscript{st} Chenyi Zhuang}
% \IEEEauthorblockA{\textit{College of Computer Science and Technology} \\
% \textit{Nanjing University of Aeronautics and Astronautics} \\
% Nanjing, China\\
% chenyi.zhuang@nuaa.edu.cn}
% \and
% \IEEEauthorblockN{1\textsuperscript{st} Pan Gao \IEEEauthorrefmark{*}}
% \IEEEauthorblockA{\textit{College of Computer Science and Technology} \\
% \textit{Nanjing University of Aeronautics and Astronautics} \\
% Nanjing, China\\
% pan.gao@nuaa.edu.cn}
% }

\author{
	 Chang Xie\textsuperscript{1*}, Chenyi Zhuang\textsuperscript{1*}, Pan Gao\textsuperscript{1}$^{\dagger}$   \\
	\textsuperscript{1}Nanjing University of Aeronautics and Astronautics, Nanjing, China \\
    \{xiechang, chenyi.zhuang, pan.gao\}@nuaa.edu.cn
}

\maketitle
\renewcommand{\thefootnote}{}
\footnotetext{* Equal contribution. $^{\dagger}$ Corresponding author.}

\begin{abstract}
Advanced diffusion models have made notable progress in text-to-image compositional generation. However, it is still a challenge for existing models to achieve text-image alignment when confronted with complex text prompts. In this work, we highlight two factors that affect this alignment: the quality of the randomly initialized noise and the reliability of the generated controlling mask. We then propose PiCo (Pick-and-Control), a novel training-free approach with two key components to tackle these two factors. First, we develop a noise selection module to assess the quality of the random noise and determine whether the noise is suitable for the target text. A fast sampling strategy is utilized to ensure efficiency in the noise selection stage. Second, we introduce a referring mask module to generate pixel-level masks and to precisely modulate the cross-attention maps. The referring mask is applied to the standard diffusion process to guide the reasonable interaction between text and image features. Extensive experiments have been conducted to verify the effectiveness of PiCo in liberating users from the tedious process of random generation and in enhancing the text-image alignment for diverse text descriptions.
\end{abstract}

\begin{IEEEkeywords}
Text-to-image diffusion model, Text-image alignment, Mask control, Random noise
\end{IEEEkeywords}

\section{Introduction}
Trained on billions of text-image pairs, Text-to-Image (T2I) diffusion models \cite{rombach2021highresolution} enable the creation of realistic images from text prompts, facilitating applications of creative expression and entertainment \cite{ramesh2022hierarchicaltextconditionalimagegeneration,li2023gligen}. The ultimate goal of T2I generation is not only to produce high-quality images but also to align with the input prompts. However, when confronted with complex text descriptions, existing T2I models still present an inconsistency between the generated image and the given text, particularly the attribute binding problem \cite{feng2023trainingfreestructureddiffusionguidance,han2023svdiff,jimenez2023mixture}. 

The first and most obvious consequence would be the need for users to do random trials and selections to obtain their ideal images. This random process could be tedious and cannot guarantee better results. What if we rank a set of noises and perform the diffusion process only on the most suitable one? Prior studies \cite{chen2023trainingfreelayoutcontrolcrossattention,xu2024good} have revealed a connection between the initial noise and the structure of the generated image. We are inspired to exploit the rich semantic information in the initial noise and develop a novel noise selection strategy to alleviate the misalignment problem caused by bad initial noise. Based on the input text, the quality of the random noise is adaptively and automatically assessed by global and concept-level scores, freeing the user from tedious manual selection. 

In addition to the initial noise, the misalignment between the textual and visual representations in the cross-attentional layers is another critical factor in causing the attribute binding issue \cite{feng2023trainingfreestructureddiffusionguidance,zhuang2024magnet}. More precisely, if desired objects are absent in the generated images, the corresponding text tokens will exhibit a low value of activation in the cross-attention maps. Similarly, the generated objects that display wrong attributes are linked to non-overlapping cross-attention activations between related text tokens. To intervene in the text-image interaction, several approaches \cite{wang2023compositionaltexttoimagesynthesisattention, liu2023detector} integrate an object detection algorithm and control attention maps through bounding boxes.
% The cross-attention mechanism where two kinds of representations interact with each other, is the best place to unveil this misalignment. For example, inactivated text tokens are associated with the problem of missing objects in the generated images. Similarly, images showing inaccurate attributes can be indicated in nonoverlapping cross-attention maps between text tokens regarding each concept. 
% To accurately convert the semantic textual information into visual representations, BoxNet \cite{wang2023compositionaltexttoimagesynthesisattention} integrates an object detection algorithm to generate bounding boxes for controlling cross-attention and self-attention maps. A concurrent work \cite{liu2023detector} directly detects each object on the attention map. 
Other works \cite{li2024get,patel2024enhancing, rassin2024linguistic} leverage the semantic information of the cross-attention maps to supervise the optimization of the image latents. However, these methods simply detect objects from the activated regions in cross-attention maps. We doubt these detection results are inaccurate and unreliable, as the text tokens can be naturally inactivated. Rather than utilizing the \textit{semantic-level} mask, we employ an off-the-shelf referring segmentation model \cite{luddecke2022image} to generate the \textit{pixel-level} mask, thereby ensuring more precise and reliable modulation.

In this work, we propose \textbf{PiCo}, a novel training-free method to enhance the text-image alignment of T2I diffusion models through two steps: \textbf{Pi}ck good noises and then \textbf{Co}ntrol attention maps. The main contributions can be summarized as:
%好像没改
% Our method enhances the alignment between generated images and text prompts by accurately correcting the attention maps of each text token. Notably, our approach is entirely training-free and does not require fine-tuning of Stable Diffusion. The main contributions of our work can be summarized as follows:
\begin{itemize}
    \item We develop a noise selection module to identify high-quality image noises that exhibit suitable representations and align with the target text at global and concept levels.
    % \item We propose \textbf{noise selection}, which leverage the pre-trained CLIPSeg model to compute ITM score for full text and Concept Score for visual concepts extract from text prompt. noise selection facilitates the identification of random seeds that align more closely with the image structures suggested by the text prompts, thereby mitigating the generation of low-matching images caused by initial random noise that is not compatible with the text prompts.
    \item We introduce a novel referring mask control module to modulate the cross-attention maps, consisting of the concept mask to improve pixel-level alignment and the exclusive mask to handle multi-concept generation.
    % \item We propose \textbf{Referring Segmentation Mask control}, which successfully employed the CLIPSeg model from the pre-trained referring segmentation task to generate attention masks for visual concepts, integrating both foreground and background control to obtain fine-grained masks for controlling the cross attention map.
    \item We have conducted various experiments to verify the proposed method, which improves synthesis quality and text-image alignment in a training-free manner.
\end{itemize}

\begin{figure*}[tb]
\centering
\includegraphics[width=\textwidth]{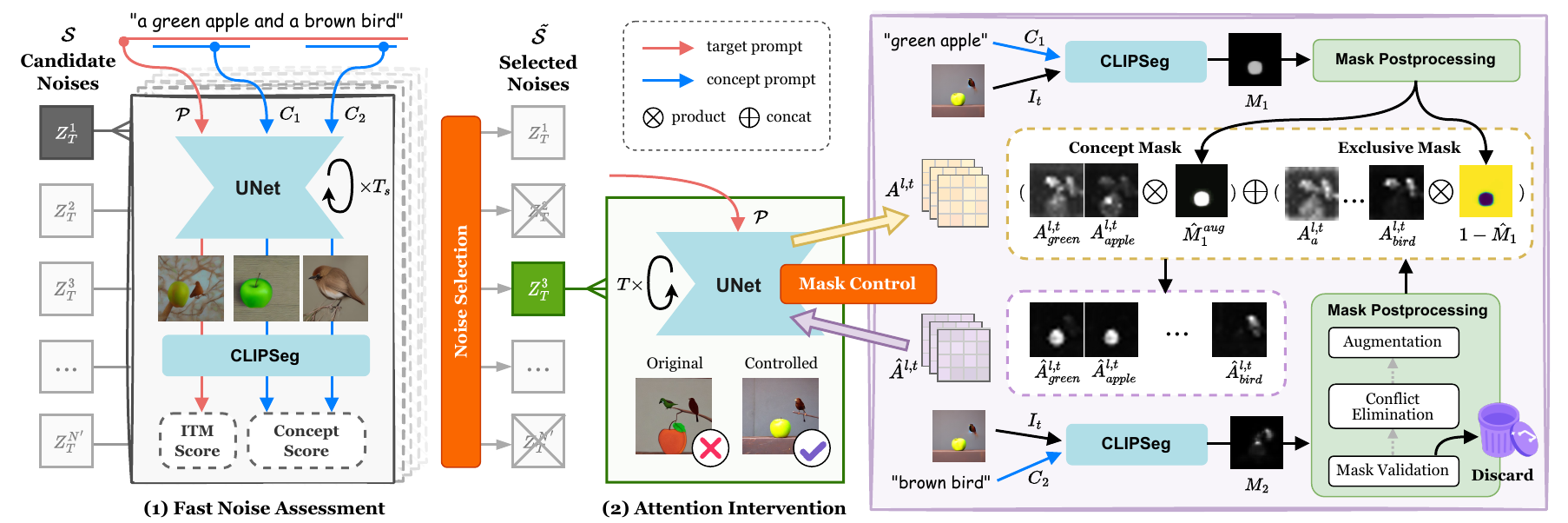}
\vspace{-6mm}
\caption{Overview of the proposed PiCo. (1) The \textit{noise selection module} fastly assesses the quality of the random noise through the ITM score and concept scores. (2) The \textit{referring mask control module} intervenes in the cross-attention layer with pixel-level concept and exclusive masks.}
\vspace{-2mm}
\label{fig:pipeline}
\end{figure*}

\section{Method}
\subsection{Overview}
As illustrated in Fig. \ref{fig:pipeline}, our proposed PiCo has two main components: (1) a \textbf{noise selection} module to select suitable initial noises based on the given text, and (2) a \textbf{referring mask control} module to perform pixel-level cross-attention manipulation. Built upon Stable Diffusion (SD) \cite{rombach2021highresolution}, PiCo improves the text-image alignment in a training-free manner.

% As illustrated in Fig. \ref{fig:pipeline}, the proposed method consists of two main components: \textbf{(1) noise selection} strategy to select the best seed that is suitable for the current text prompt from a set of random seeds to generate initial random noise. In our denoising process, denotes as \textbf{(2) Referring Segmentation Mask control}, we use the decoded intermediate image $I_t$ obtained in time step $t$ to generate masks for all the visual concepts in the text prompt analyzed by NLP tokenizer using the CLIPSeg model. After mask postprocessing module, we obtain a set of masks, which are then applied to the corresponding cross attention map. By optimizing the alignment between text prompts and their corresponding cross-attention maps, we can enhance the quality of generated images and improve the accuracy of image-text matching.

\subsection{Fast Noise Selection}
\label{Sec:seed_selection}
The initial noise is crucial for text-to-image generation in determining the image structure \cite{xu2024good}. Intuitively, a good noise allows the attendance of all concepts in the given prompt. Note that this could be \textit{prompt-independent}, i.e., one noise that is good for the current prompt may be bad for another. Yet, using the generated images to check the validity of the noise would be time-consuming. In this work, we try to answer this question: \textbf{Can we efficiently assess the quality of noises for specific prompts?} It is said that the same initial noise leads to similar high-level features in different sampling steps \cite{song2020denoising}. We are motivated to design a prompt-aware and fast noise selection module. The efficiency is ensured by setting a small denoising step $T_s$ (e.g., $T_s=5$ against standard $50$ steps). 

To assess the quality of each noise, we propose to compute a global-level score from the full prompt and concept-level scores from separated visual concepts. Consider generating $N$ images for a given prompt $P$, we construct a candidate set with $N'$ unique noises, denoted as $\mathcal{S}=\{z_T^1,...,z_T^{N'}\}$. The size of the candidate set is determined by a pre-defined ratio $r_s$ as $N'=N\cdot r_s$. To compute the global-level \textit{ITM} (Image-Text Matching) score for each initial noise, we perform $T_s$ denoising steps conditioned on the full prompt $P$ and obtain the image $I_P$. The ITM score is calculated by:
\begin{equation}
    ITM\;Score = cos(F_1(I_P), F_2(P)),
\end{equation}
where $F_1$, $F_2$ are the image and text encoders taken from the pre-trained CLIPSeg \cite{luddecke2022image}, respectively. Note that this generated image $I_{P}$ is coarse due to insufficient sampling steps. However, the ITM score is intended to measure the agreement between the structural information and the target text. This fast sampling results in inferior image quality, but provides enough semantics to achieve this goal.

The compositional generation has been observed with the object neglect issue due to entangled textual representations \cite{chefer2023attendandexciteattentionbasedsemanticguidance, feng2023trainingfreestructureddiffusionguidance}. We are inspired to associate the noise quality with single-concept generation: if the initial noise does not produce a meaningful result, it must not be suitable for full prompt generation in a more complex context. To compute these concept-level scores, we use Stanza's dependency parsing module \cite{qi2020stanza} to obtain $R$ objects and their related attributes, denoted as $\mathcal{C}=\{C_1,...,C_R\}$. Each concept is used to condition the generation, resulting in $R$ concept images as $I_{r},r=1,...R$. Next, we utilize CLIPSeg to produce the segmentation result $M_{r}$ from $I_{r}$ based on the text query $C_r$. The concept score is measured from two perspectives: aggregation and intensity. The aggregation degree is defined as the averaged value of elements greater than a threshold $\tau_{r}$, which is adaptively determined by the $90$-th percentile of the activation values in $M_{r}$. We formulate the above process as:
\begin{equation}
    v^{avg}_{r} = \frac{1}{H}\sum_{h=1}^{H}{x_h}, \quad where \quad  x_h > \tau_{r} \;and\; x_h \in M_{r},
\end{equation}
where $H$ represents the number of image pixels that have greater value than $\tau_r$. To assess the intensity, we focus on the max value of the segmentation as:
\begin{equation}
v^{max}_{r} = max(M_{r}),
\end{equation}
The aggregation degree $v^{avg}_{r}$ assesses the region of the visual concept that is present in the image $I_{r}$, while the intensity degree $v^{max}_{r}$ highlights the significance of the concept $C_r$ appearing in $I_{r}$. We suspect that the latter is more crucial for the attendance of the concept. We assign a larger weight to the max value $v^{max}$ and normalized the mean value $v^{avg}$ on the max value to stress the intensity of all visual concepts. The concept score is defined as:
\begin{equation}
    Concept\;Score = v^{max} \cdot \delta + \frac{v^{avg}}{v^{max}},
\label{eq:concept_score}
\end{equation}
where $\delta$ is a hyperparameter. Finally, the overall noise score is the sum of the ITM score and $R$ concept scores. 

Note that we employ the referring segmentation model rather than traditional object detection models such as YOLO\cite{redmon2016you}, since the former allows arbitrary referents while the latter only supports pre-defined classes. The noise selection generated $R+1$ images in total, while it is guaranteed to be efficient and time-saving by the small steps $T_s$ and parallel computing (see Tab. \ref{tab:Ts_ablation}). We select the best $N$ noises with the highest scores in the candidate set $\mathcal{S}$. We denote the selected noises as $\mathcal{\tilde{S}}$. In the following, we introduce our cross-attention controlling strategy based on each initial noise $z_T$ in $\mathcal{\tilde{S}}$.

\subsection{Referring Mask Control}
The suitable noise $z_T\in \mathcal{\tilde{S}}$ increases the chances of generating a faithful image under a standard $T$-step denoising. However, cross-attention between text and image features also raises the issue of misalignment. Driven by this, we then intervene in the denoising U-Net and introduce the referring mask control module. Other than previous works \cite{liu2023detector} that extract \textit{semantic-level} masks from cross-attention maps, we adopt CLIPSeg \cite{luddecke2022image} to produce \textit{pixel-level} masks for more precise manipulation. 

We propose to apply the referring mask control at early denoising stages $t\in[T, T_{c}], 1\leq T_c\leq T$, where the visual representation are more controllable \cite{chefer2023attendandexciteattentionbasedsemanticguidance}. We denote the intermediate noise latent as $z_t=\epsilon_\theta(z_{t+1}, t+1, P)$ and the corresponding cross-attention maps at $l$-th layer as $\mathcal{A}^{l,t}=\{A_{SOT}^{l,t}, A_1^{l,t},..., A_p^{l,t},A_{EOT}^{l,t}\}$, where $SOT,EOT$ are two special symbols, $p$ is the text length. We project this noisy image from the latent space into the image space by $I_t=\mathcal{D}(z_t)$, where $\mathcal{D}$ is the SD decoder. Next, we prompt the CLIPSeg model with referents of the intermediate image $I_t$ and each concept text $\{C_1,...,C_R\}$, resulting in referring segmentations $\mathcal{M}=\{M_1,...,M_R\}$. Intuitively, the segmentation $M_r\in \mathcal{M}$  be directly converted into a binary mask to control the cross-attention computation. However, we find in practice that $I_t$ at the early denoising stage is blurry and vague. In this case, the extracted masks have sparse highlights on background regions, which are not reliable for controlling cross-attention. Therefore, we further rectify these coarse segmentations into precise referring masks. To determine whether a segmentation $M_r$ is valid, we focus on elements that have greater values than a threshold $\rho_r$. This threshold $\rho_r$ is adaptively computed by the $90$-th percentile of activation. Next, we calculate the standard deviation (std) of their locations from the x- and y-coordinates:
\begin{equation}
\sigma^{x}_r,\sigma^{y}_r=std(x,y\in\mathop{\arg\max}\limits_{x\in w,y\in h} M_r(x,y) > \rho_r),
\end{equation}
where $h,w$ are the height and width. The dispersion of these high-value elements of the $r$-th concept mask is defined as:
% \begin{equation}
% \sigma^{x,i}_r=std(topK(M_r(x,i), \rho)), i\in h
% \end{equation}
% \begin{equation}
% \sigma^{y,j}_r=std(topK(M_r(j,y), \rho)), j\in w
% \end{equation}

% \begin{equation}
%     coord = \mathop{\arg\max}\limits_{\rho}(x,y|M_r(x,y), x\in w, y\in h)
% \end{equation}
% To measure the aggregation degree, we use the standard deviation (std) of the x-coordinate as $x_{std}=std(M_r(x,i))$ and that of the y-coordinate as $y_{std}=std(M_r(j,y))$ to calculate:
\begin{equation}
    \sigma_r = \sigma^{x}_r\cdot \sigma^{y}_r,
\end{equation}
We also leverage the maximum value of activation to avoid the low-confidence risk. The validity of each segmentation is assessed as:
\begin{equation}
    \hat{M}_r=\left\{
    \begin{aligned}
        &\text{discard} , & if\;max(M_r)<\alpha_{l}\;and\;\sigma_r>\alpha_{h}\\
        &\text{Sigmoid}(M_r) , & otherwise
    \end{aligned}
    \right.
    ~,
\end{equation}
where $\alpha_{l},\alpha_{h}$ are two hyperparameters of thresholds. If the segmentation $M_r$ is sparse (i.e., $\sigma_r>\alpha_{h}$) and has small logits (i.e., $max(M_r)<\alpha_{l}$), it is considered unreliable and will be discarded. To maintain the semantic information, we do not intervene in the related cross-attention maps.

\textbf{Conflict Elimination}.  When generating multiple concepts, overlapped activations between different concepts can lead to the sticking issue. In other words, the current pixel of visual features interacting with textual features from multiple concepts produces entangled representation. To prevent multiple-concept overlapping, we assign conflicted pixel to the mask with the highest value. For each pixel $(x,y)$ in the conflict area  under $R'$ visual concepts as $\{\hat{C}_{1},...,\hat{C}_{R'}\}$, we obtain the index map to indicate which visual concept to occupy the pixel:

\begin{equation}
    U(x,y) = \mathop{\arg\max}\limits_{r'=1,...,R'} (\hat{M}_{r'}(x,y))
\end{equation}
We eliminate the conflict on all concept masks by: 
\begin{equation}
\hat{M}_{r}(x,y) = \mathds{1}(U(x,y)=r) \odot \hat{M}_{r}(x,y)
\end{equation}
% Even though segmentation results are more precise than bounding boxes, we still encounter some overlapping conflicts in practice. We determine the attribution of each pixel in the conflict area between all masks pairwise based on its value on the conflict mask:

% \begin{equation}
%     \mathop{\arg\max}\limits_{g\neq r}(\hat{M}_r(x,y),\hat{M}_g(x,y))
% \end{equation}

% \begin{equation}
%     \left\{
%     \begin{aligned}
%         &\hat{M}_g(x,y)=0, &\hat{M}_r(x,y)>\hat{M}_g(x,y) \\
%         &\hat{M}_r(x,y)=0, &\hat{M}_r(x,y)<\hat{M}_g(x,y) \\ 
%         &\hat{M}_r(x,y)=\hat{M}_g(x,y)=0, &\hat{M}_r(x,y)=\hat{M}_g(x,y) \\ 
%     \end{aligned}
%     \;
%     \right.
%     ~,
% \end{equation}
% where $g\neq r$ and $M_{conflict}$ represents the conflict area between two masks. We keep the original value of the mask with a higher value in pixel $x,y$ and assign zero to the conflict mask to eliminate those conflict areas.
% The conflict area we define as follows, $\rho$ is a hyperparameter:
% \begin{equation}
%     \text{Conflicts}=\text{argwhere}((\hat{M}_r(x,y)>\rho)~~\&~~(\hat{M}_s(x,y)>\rho))
% \end{equation}

\textbf{Augmented Concept Mask}. In practice, even with the referring mask $\hat{M}_r$, insufficient activations in $\mathcal{A}^{l,t}$ cannot encourage the concept to appear in the final image. Meanwhile, the zero region of these masks eliminates the semantic information, yet it is important in the early stages for maintaining a relatively blurred and smooth state. In this case, we amplify significant regions and suppress irrelevant regions:
\begin{equation}
    \hat{M}^{aug}_r = \left\{
    \begin{aligned}
        &\hat{M_r}(x,y)\cdot \gamma, &\hat{M}_r(x,y)\geq \beta_l \\
        &\frac{\hat{M_r}(x,y)}{\gamma}, &\hat{M}_r(x,y)\leq \beta_h \\ 
    \end{aligned}
    \;
    \right.
    ~,
\label{eq:augmentation}
\end{equation}
where $\beta_l$ and $\beta_h$ are lower and upper thresholds for this mask augmentation, $\gamma$ is a hyperparameter of strength. 

Assume the current visual concept $C_r$ is represented by text tokens $a:b$, where $1\leq a\leq b \leq p$. We apply this augmented concept mask $\hat{M}^{aug}_r$ to the corresponding cross-attention maps $A_{a:b}^{l,t}=\{A_a^{l,t},...A_b^{l,t}\}$ to perform the referring mask control:
\begin{equation}
    A_{a:b}^{l,t}:=A_{a:b}^{l,t} \odot \hat{M}^{aug}_r ,
\end{equation}

\textbf{Exclusive Mask for Multi-concept}. we invert all concept masks that do not belong to the current concept as the exclusive mask to control the updated cross-attention maps:
\begin{equation}
    A_{a:b}^{l,t}:=A_{a:b}^{l,t} \odot \prod_{g\neq r}(1 - \hat{M}_g),
\end{equation}
For other text tokens like conjunctions, we also employ the exclusive mask by reversing all concept masks:
\begin{equation}
    A_{q}^{l,t}:=A_{q}^{l,t} \odot \prod^R_{r=1}(1 - \hat{M}_r),
\end{equation}
where $1\leq q \leq p$, $\odot$ is the element-wise multiplication.

\begin{figure*}[tb]
\centering
\includegraphics[width=0.9\textwidth]{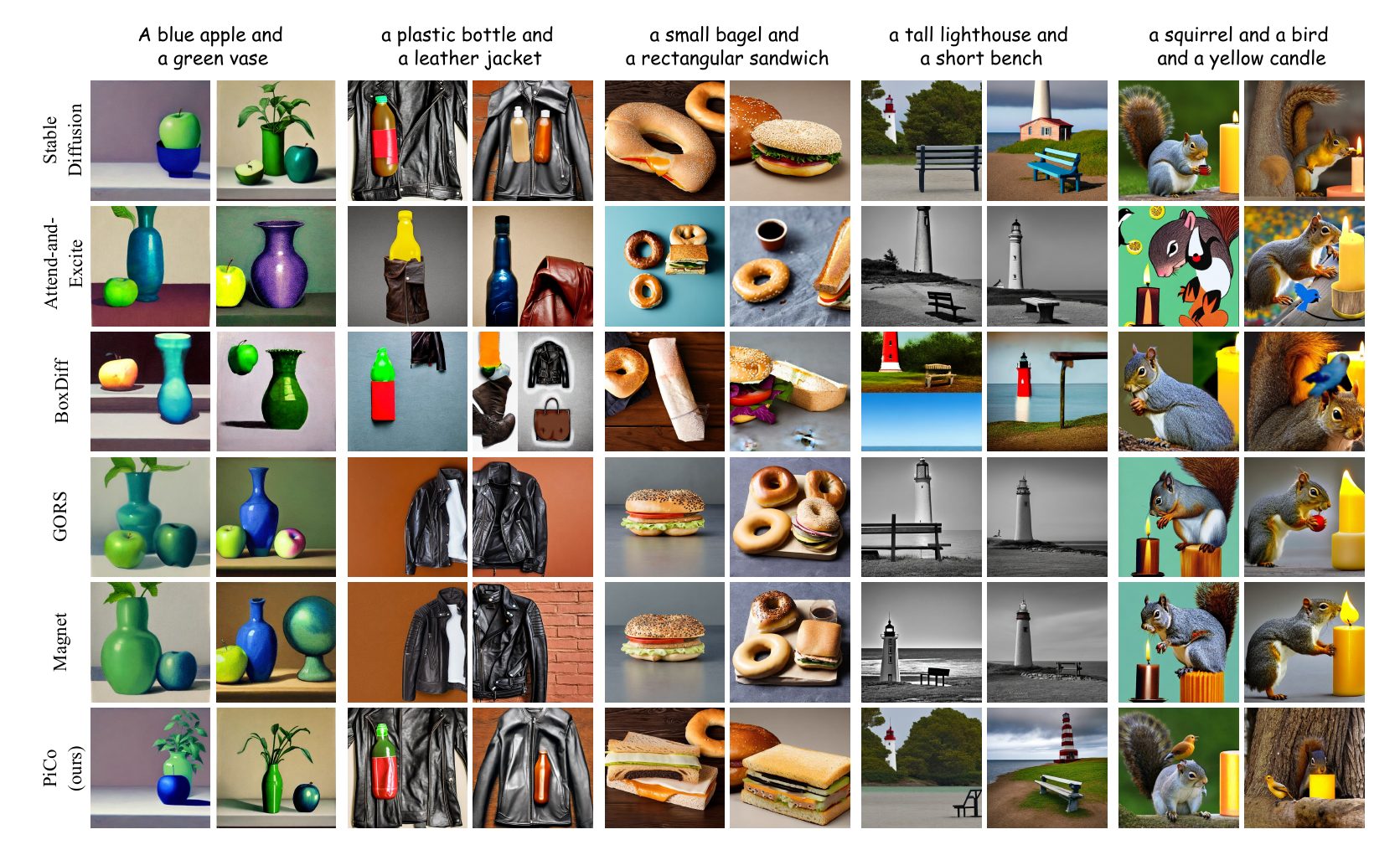}
\vspace{-2mm}
\caption{Qualitative comparison using prompts from T2I-Compbench. Each case uses the same initialized noise, which is output by the noise selection module. }
\vspace{-2mm}
\label{fig:qualitative}
\end{figure*}

\section{Experiments}
\subsection{Experimental setup}
\textbf{Implementation Details}. We implement PiCo on SD v2.1. For the noise selection module, we perform 5 DDIM steps (i.e., $T_s=5$) and set the ratio $r_s=10$ for the candidate set size, $\delta=2$ for score computation. For the referring mask control module, we intervene in early denoising stages $T_c=25$ from the total of 50 DDIM steps (i.e., $T=50$) to generate the final image. We set $\alpha_l=0.7,\alpha_h=1500$ for validation assessment, $\beta_l=0.5,\beta_h=0.7$, and strength $\gamma=15$ for augmentation. 
% \textbf{Hyperparameters}. We evaluate the proposed method on SD v2.1. For the noise selection module, we adopt the set $T_s=5$ and the seed set size by $Q=10$. For the referring mask control module, we perform $T=50$ DDIM steps with $T_c=25$ for the mask controlling. The augmentation strength is set to $\gamma=15$. More hyperparameters setting please refer to Appendix C.

% The classifier-free guidance scale is set to $7.5$.

% We adopt the DDIM scheduler with 5 denoising steps ($T_s=5$) for noise selection module, and we set the number of seeds $N$ in the random seed set to 10. %while the number we choose for selected seeds is $k=3$. 
% We set $\lambda = 90$ and $\delta = 2$. In Referring Segmentation Mask control module, the DDIM scheduler with 50 denoising steps ($T=50$), the number of controlling steps $T_c$ is set to 25 as default. When calculating standard deviation, we set $\rho=90\%\cdot (hw)$. In generating $\hat{M}_s$, we set $\alpha=0.7$ and $\beta=1500$, further more, we set $p_u = 0.7$ and $p_l=0.3$, the enhancement and suppression strength $\sigma$ is set to 15. The ratio of classifier-free guidance is set to 7.5. The version of Stable Difsion we choose is SD v2.1 .

\textbf{Datasets}. We evaluate PiCo on two benchmarks: CC-500\cite{feng2023trainingfreestructureddiffusionguidance}, T2I-Compbench \cite{huang2023t2i}. We randomly select 100 prompts in CC-500. T2I-Compbench consists of 6,000 prompts, grouped into 3 categories for attribute binding (three sub-categories: color, shape, and texture), object relationships, and complex compositions. We select 100 prompts for each attribute binding subcategories (color, shape, and texture) and 100 prompts from complex compositions. During the evaluation, we randomly selected 500 prompts and 10 unique seeds per prompt (generated 5,000 images in total). Considering the human cost, we randomly sampled 500 images with two objects for the subjective evaluation. For objective metrics, we assessed PiCo on 4,000 images to obtain a reliable result.

\begin{table*}[tb]
\caption{Annotated comparison of human evaluation and objective metrics for attributes binding. We assign the selected noises from PiCo to all methods for a fair comparison.}
\centering
\rowcolors{6}{gray!10}{gray!10}
\begin{tabular}{ccccccccccc}
\toprule
     \multirow{2}{*}{Model}&\multicolumn{3}{c}{Random}&\multicolumn{3}{c}{Selected}&\multicolumn{4}{c}{Objective Metrics (Selected)}\\
    \cmidrule(r){2-4}\cmidrule(r){5-7}\cmidrule(r){8-11}&Both($\uparrow$)&None($\downarrow$)&DINO($\uparrow$)&Both($\uparrow$)&None($\downarrow$)&DINO($\uparrow$)&CLIP-I($\uparrow$)&FID($\downarrow$)&VQAScore($\uparrow$)&CLIP-S($\uparrow$)\\
    \hline
    Stable Diffusion&27.5&34.0&0.544&34.7(\textcolor{red}{+6.8})&19.8(\textcolor{blue}{-14.2})&0.659(\textcolor{orange}{+0.115})&70.73&41.65&0.45&16.24\\
    %\rowcolor{gray!10}
    Attend-and-Excite&31.5&27.7&0.628&38.0(\textcolor{red}{+6.5})&14.5(\textcolor{blue}{-13.2})&0.723(\textcolor{orange}{+0.095})&71.47&38.24&0.47&16.35\\
    %\rowcolor{gray!10}
    BoxDiff&29.5&31.6&0.564&35.7(\textcolor{red}{+6.2})&19.3(\textcolor{blue}{-12.3})&0.680(\textcolor{orange}{+0.116})&68.55&43.59&0.42&16.22\\
    %\rowcolor{gray!10}
    GORS&32.0&28.3&0.615&39.5(\textcolor{red}{+7.5})&14.0(\textcolor{blue}{-14.3})&0.707(\textcolor{orange}{+0.092})&71.51&36.60&0.45&16.32\\
    %\rowcolor{gray!10}
    Magnet&32.5&27.7&0.623&38.3(\textcolor{red}{+5.8})&15.0(\textcolor{blue}{-12.7})&0.716(\textcolor{orange}{+0.093})&72.34&36.08&0.43&16.29\\
    \rowcolor{gray!50}
    PiCo (Ours)&36.0&23.5&0.637&41.8(\textcolor{red}{+5.8})&11.8(\textcolor{blue}{-11.7})&0.730(\textcolor{orange}{+0.093})&71.63&35.74&0.50&16.26\\
    \rowcolor{white}
\bottomrule
\label{tab:attr}
\end{tabular}
\end{table*}

% \begin{table*}[tb]
% \caption{Automatic benchmarks comparison and human evaluation for text-image alignment, here we assign our noise selection to all methods for fair comparison.}
% \centering
% \begin{tabular}{ccccccccc}
% \toprule
%      \multirow{2}{*}{Model}&\multicolumn{2}{c}{Automatic}&\multicolumn{3}{c}{Attr. (w/o SS)}&\multicolumn{3}{c}{Attr. (w/ SS)}\\
%     \cmidrule(r){4-6}\cmidrule(r){7-9}\cmidrule(r){2-3}&VQAScore&CLIPScore&Both($\uparrow$)& Single&None($\downarrow$)&Both($\uparrow$)& Single&None($\downarrow$)\\
%     \hline
%     Stable Diffusion&a&15.30&27.5&38.5&34.0&34.7(\textcolor{red}{+6.8})&45.5&19.8(\textcolor{blue}{-14.2})\\
%     Attend-and-Excite&a&b&31.5&40.8&27.7&38.0(\textcolor{red}{+6.5})&47.5&14.5(\textcolor{blue}{-13.2})\\
%     BoxDiff&a&b&29.5&38.9&31.6&35.7(\textcolor{red}{+6.2})&45.0&19.3(\textcolor{blue}{-12.3})\\
%     GORS&a&b&32.0&39.7&28.3&39.5(\textcolor{red}{+7.5})&46.5&14.0(\textcolor{blue}{-14.3})\\
%     Magnet&a&b&32.5&39.8&27.7&38.3(\textcolor{red}{+5.8})&46.7&15.0(\textcolor{blue}{-12.7})\\
%     Ours&a&16.69&36.0&40.5&23.5&41.8(\textcolor{red}{+5.8})&46.3&11.8(\textcolor{blue}{-11.7})\\
% \bottomrule
% \label{table_auto}
% \end{tabular}
% \end{table*}

\begin{table}[tb]
\caption{General comparison of human evaluation in terms of image quality and text-image alignment.}
\centering
\rowcolors{5}{gray!10}{gray!10}
\begin{tabular}{ccccccc}
\toprule
    PiCo (Ours)&\multicolumn{3}{c}{Quality}&\multicolumn{3}{c}{Alignment}\\
    \cmidrule(r){2-4}\cmidrule(r){5-7}v.s.&Win($\uparrow$) &Loss&Tie&Win($\uparrow$)&Loss&Tie\\
    \hline
    Stable Diffusion&\cellcolor{gray!50}\textbf{25.8}&19.2&55.0&\cellcolor{gray!50}\textbf{51.7}&20.8&27.5\\
    Attend-and-Excite&\cellcolor{gray!50}\textbf{42.5}&25.8&31.7&\cellcolor{gray!50}\textbf{43.3}&19.2&37.5\\
    BoxDiff&\cellcolor{gray!50}\textbf{44.2}&17.5&38.3&\cellcolor{gray!50}\textbf{45.8}&16.7&38.2\\
    GORS&\cellcolor{gray!50}\textbf{34.4}&30.0&35.6&\cellcolor{gray!50}\textbf{40.8}&24.2&35.0\\
    Magnet&\cellcolor{gray!50}\textbf{34.2}&30.8&35&\cellcolor{gray!50}\textbf{41.7}&21.6&36.7\\
    \rowcolor{white}
\bottomrule
\label{tab:compare}
\end{tabular}
\end{table}

\textbf{Baselines}. We compare PiCo to the SD v2.1 \cite{rombach2021highresolution}, Magnet~\cite{zhuang2024magnet} which is training-free, Attend-and-Excite \cite{chefer2023attendandexciteattentionbasedsemanticguidance} and BoxDiff \cite{xie2023boxdiff} which optimize the noise latent, and GORS \cite{huang2023t2i} which fine-tunes the denoising U-Net. We adopt the official implementation and default hyperparameters for all baselines. 
% We utilize GPT-4o \cite{OpenAI-GPT-4o} to provide the layout control for the additional input of BoxDiff (see Appendix C).

\textbf{Metrics}. Following previous works \cite{zhuang2024magnet, feng2023trainingfreestructureddiffusionguidance}, we rely on human evaluation as automatic metrics are unreliable in assessing attribute binding. Given the novelty of assessing the noise quality, there is no \textit{fair} baselines to compare our noise selection module with. Therefore, we input all compared methods with random noises and selected noises and validate the proposed PiCo from the following perspectives: (1) \textbf{Annotated Comparison}. Annotators are asked to label the number of correct concepts. For example, given the prompt ``a blue apple and a green vase", annotators indicate one image for two (any two concepts appear), one (only one concept) or zero (none of them). 
% Notice that concepts with explicit attributes but present inconsistent attributes in the generated images will not be counted (e.g., ``red candle"). 
All baselines use random noises (i.e., without noise selection), and PiCo's selected noises (i.e., with noise selection) to generate images, respectively. The effectiveness of the noise selection module can be verified if the latter setting shows improvement against all methods. (2) \textbf{General Comparison}. We ask human evaluators to compare two images (one is ours and another from baselines) based on two metrics: \textit{image quality} for more visually appealing and \textit{text alignment} for more faithful to target. If two images are equally good or bad, evaluators can choose ``Tie". During this comparison, all methods are controlled under the selected noises, therefore we prove the effectiveness of the proposed referring mask control module if ours can outperform others. (3) \textbf{Objective Metrics}. For objective metrics, we evaluated our PiCo on 4000 images (400 prompts from T2ICompbench dataset) with CLIPScore, VQAScore~\cite{chefer2023attendandexciteattentionbasedsemanticguidance} and DINO-Score. For CLIP-I and FID, we tested these two metrics on the MS-COCO~\cite{lin2014microsoft} dataset. More discussion of these metrics can be found in Appendix B.
% The human evaluation was primarily conducted on the CC-500 and T2ICompbench. Given the prompt ``a white cat and a gray mouse", each annotator is required to provide a response corresponding to the associated image. For attribute binding, annotators evaluate the number of objects in the image whose attributes match the text prompt, indicating two (both objects appear in the image), one (``cat" or ``mouse" appears), or zero (none of them). For comparison, annotators need to choose the better result based on image quality and alignment between two images (from two methods), and if they are the same, it is considered a tie.
%We also report the Fréchet Inception Distance (FID) score computed on 5,000 images on the MSCOCO dataset, as well as image-text matching metrics VQAScore and CLIPScore conducted on the T2I-Compbench dataset.
% \textbf{Automatic Evaluation.} To ensure a fair comparison with other methods, we employed automated evaluation metrics. Specifically, we utilized the FID to assess image quality on the MSCOCO dataset. For image-text matching evaluation, we applied metrics VQAScore and CLIPScore. These automatic metrics are mainly conducted on the 400 text prompts we selected from the T2I-Compbench dataset.
\subsection{Comparison Results}

\textbf{Quantitative results}. Tab. \ref{tab:attr} compares the generation under random initialized noises or selected noises, i.e., without or with the proposed noise selection module. With the selected noises, the generated images of all methods have a significant improvement in the ``Both" rate (marked in red) and the DINO score (marked in orange), and a remarkable decrease in the ``None" rate (marked in blue). This comparison shows the effectiveness of our proposed noise selection module in alleviating the problem of object neglect. Our noise selection module can filter out bad initial noises and increase the odds of visual concept attendance. Note that in both settings, PiCo outperforms baselines for encouraging the most objects (highest in ``Both" and lowest in ``None"). We also report other commonly used objective metrics, where our PiCo outperforms baselines in terms of FID and VQAScore. Tab. \ref{tab:compare} presents the general comparison of human evaluation. All cases in this comparison are controlled by selected noises. In both terms of image quality and text alignment, PiCo gets more votes than baselines. Annotators indicate more than a 25\% chance of our method winning the comparison in the quality, and a 40\% chance in the alignment. This verifies the effectiveness of our proposed referring mask module.

\textbf{Qualitative results}. Fig. \ref{fig:qualitative} presents results from the T2ICompbench dataset. Attend-and-Excite and BoxDiff optimize the image latent to ensure the attendance of objects while showing the sticking issue (e.g., ``bottle" and ``jacket" are blended in column 3) or inaccurate attributes (e.g., ``leather bottle" rather than ``plastic bottle" in column 4). In addition, their results have unwanted artifacts that significantly affect the synthesis quality (e.g., unrealistic ``bird" in columns 9-10). Interestingly, the fine-tuning method GORS and the training-free method Magnet present similar results and still suffer from severe missing object problem (e.g., fail to generate ``bottle" in columns 3-4 and ``bird" in columns 9-10). Conversely, PiCo generates visually appealing images with multiple concepts and accurate attributes. Notice that PiCo can handle unnatural concepts such as ``rectangular sandwich", for which baselines produce ambiguous results. This comparison indicates that our \textit{pixel-level} mask provides precise guidance for control of object shape, color, and texture.

\begin{table}[tb]
\caption{Ablation studies for the inference step $T_s$ in the noise selection module.}
\centering
\rowcolors{4}{gray!50}{gray!10}
\begin{tabular}{lcccc}
\toprule
     \multirow{2}{*}{$T_s$}&\multicolumn{3}{c}{Human Annotation}&Runtime(s)\\
     \cmidrule(r){2-4}
     &Both($\uparrow$)&Single&None($\downarrow$)&Per Prompt\\
\midrule
    1& 28.2 &38.9 & 32.9& 0.84\\
    \rowcolor{gray!50}
    5 (Ours)& 36.2 &45.4 & 18.4& 2.03\\
    20& 51.0&44.7 & 4.3 & 6.41\\
    \rowcolor{white}
\bottomrule
\label{tab:Ts_ablation}
\end{tabular}
\end{table}

\subsection{Ablation Study}
\textbf{Noise selection Inference Steps $T_s$}. To accelerate the noise selection, we propose to use small sampling steps. Tab.  \ref{tab:Ts_ablation} tests three settings for $T_s=1,5,20$, respectively. When setting $T_s=1$, the generated images maintain a noisy state (see Appendix F) and cannot provide reliable scores, resulting in the worst performance as its highest rate of ``None". Conversely, when setting $T_s=20$, the rate of ``Both" reaching 51\% indicates the best performance. More steps of denoising lead to meaningful noise scores, yet significantly increase the time resource. We empirically set $T_s=5$ for a balance between time and performance. 

\begin{figure}[tb]
\centering
\includegraphics[width=\linewidth]{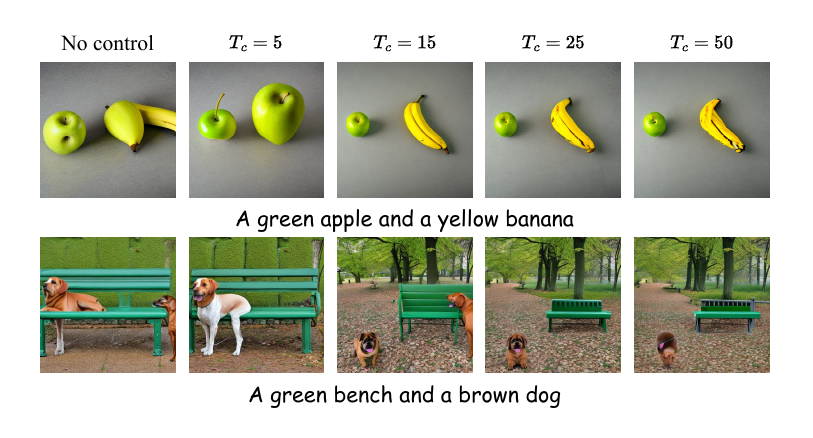}
\vspace{-8mm}
\caption{Ablation study on $T_c$ to stop the mask control. Under a small value, the intervention is not enough but a larger one can lead to degeneration. }
\label{fig:control_step}
\end{figure}

\textbf{Mask Control Steps $T_c$}. Fig. \ref{fig:control_step} studies the hyperparameter $T_c$ to perform the referring mask control. To generate images under $50$ DDIM steps, we validate the decision to stop the mask control at $T_c=5,15,25,50$. Compared to no control, the result of $T_c=5$ shows limited improvement of matching between text and images due to insufficient intervention effect. Similarly, $T_c=15$ encourages the object to some extent while still not satisfactory. While using full mask control as $T_c=50$, the interference is too strong to produce high-quality images. We find that stopping at half of all denoising steps i.e., $T_c=25$, provides a suitable intervention effect without significantly affecting the synthesis quality. 

\begin{figure}[tb]
\centering
\includegraphics[width=\linewidth]{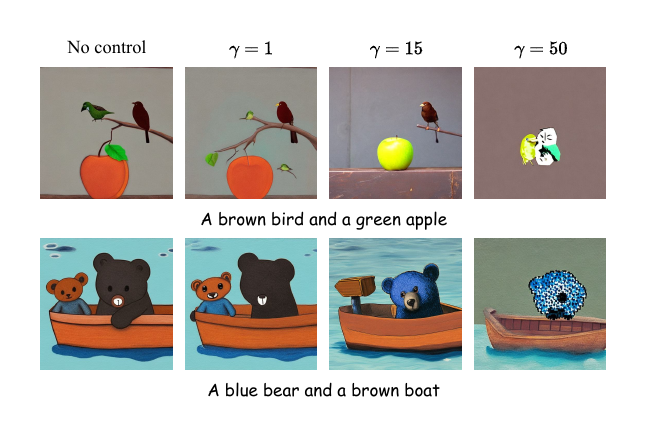}
\vspace{-10mm}
\caption{Ablation study on $\gamma$. A small value of $\gamma$ can not well disentangle different concepts, while a large value causes artifacts in the generated image (best viewed zoomed in). We empirically set $\gamma = 15$.}
\label{fig:gamma}
\end{figure}

\textbf{Augmentation Strength $\gamma$.} As shown in Fig. \ref{fig:gamma}, we study the value of $\gamma$ in Eq. (\ref{eq:augmentation}) during the mask augmentation. Observe that $\gamma=1$ generates a similar result to the vanilla model, indicating an inadequate augmentation. On the other hand, the modulation under $\gamma=50$ has a similar phenomenon to the mask control steps under $T_c=50$ that excessive intervention effect causes the collapse of image quality. 
% For example, ``blue bear" in the last column presents out-of-distribution image representations and unrealistic artifacts. 
We empirically set $\gamma=15$ which performs well in most cases.

\textbf{More Ablation Studies}. We have also conducted extensive ablation studies to verify our proposed method and choice of hyperparameters. Please refer to Appendix D for more details.

% \begin{figure}[tb]
% \centering
% \includegraphics[width=0.9\columnwidth]{Picture/inter-intra.pdf}
% \caption{Ablation study. We demonstrated the necessity of implementing two types of mask control.}
% \label{fig:inter-intra}
% \end{figure}
% \subsection{Extensions}
% 还没改
\section{Conclusion}
In this work, we introduce an innovative training-free approach to improve the text-image alignment of T2I diffusion models. Since the initial noise plays a crucial role in the image layout, we are inspired to design a noise selection module to provide reliable noises for generating high-quality and faithful images. To further enhance the alignment, we propose the referring mask control module to precisely manipulate the cross-attention maps during the denoising process. Extensive experiments show that our proposed PiCo achieves better alignment between the generated image and the target text. The noise selection strategy is feasible to alleviate various T2I issues such as missing objects. These pixel-level masks can precisely modify the visual representation with a slight computational cost. We hope that this paper can raise awareness of the noise quality and provide novel insights for more accurate manipulation of the cross-attention module.

\bibliographystyle{IEEEbib}
\bibliography{main}

\appendix
\subsection{Preliminaries}
\textbf{Latent Diffusion Model (LDM)}. Exemplified by Stable Diffusion (SD) \cite{rombach2021highresolution}, LDM performs the diffusion process in the latent space. The workflow of the diffusion model is usually divided into two stages: (1) the forward diffusion iteratively adds Gaussian noise to the input image $x_0$ at each timestep $t$, and (2) the reverse process employs a diffusion model to predict the entire noise to be removed from the noisy image $x_t$. SD trains an autoencoder to project the image $x$ into a spatial latent code $z=\mathcal{E}(x)$ and reconstruct via $\hat{x} = \mathcal{D}(\mathcal{E}(x))$, where $\mathcal{E}$ and $\mathcal{D}$ are the image encoder and decoder, respectively. Therefore, the above forward and reverse process is performed on the noisy latent $z_t$ at a specific timestep $t$. To condition the generation with additional text $y$, SD adopts the pre-trained CLIP \cite{radford2021learning} text encoder to obtain text embeddings $\mathcal{E}_{text}(y)$ and appends several cross-attention layers to inject the text condition into the latent $z_t$. The denoising network $\epsilon_{\theta}$ is trained to minimize $||\epsilon_{\theta}(z_t,t,y)-\epsilon||^2$, i.e., removing the added Gaussian noise $\epsilon\sim \mathcal{N}(0,1)$ at timestep $t$ with textual condition $y$. At inference, a noisy latent $z_T$ is sampled from the Gaussian distribution $\mathcal{N}(0,1)$, and is iteratively denoised to the final latent $z_0$ using the trained denoising model $\epsilon_{\theta}$ to output the image $x$ through $x=\mathcal{D}(z_0)$. 

% Given an image $x$, the encoder $\mathcal{E}$ maps it to a latent $z$, and the decoder $\mathcal{D}$ reconstructs the image from the latent, i.e., $\overline{x} = \mathcal{D}(z)=\mathcal{D}(\mathcal{E}(x))$. At each timestep $t$ in forward process, a noisy latent $z_t$ is obtained, and the model iteratively refines this representation to generate the final image. And a denoising network $\epsilon_{\theta}$ is trained to minimize $||\epsilon_{\theta}(z_t,t)-\epsilon||^2$ for removing the added noise at timestep $t$, where $z_t$ is the noisy latent at timestep $t$, $\epsilon\sim \mathcal{N}(0,1)$ is the added Gaussian noise. 

% Stable Diffusion introduces conditioning mechanisms to control the synthesized image content through an additional input, such as a text prompt $y$. The text prompt is processed into text embedding $\tau_{\theta}(y)$ by pretrained CLIP \cite{radford2021learning} text encoder. A conventional DDPM\cite{ho2020denoising} model can be trained using the following formula:
% \begin{equation}
%     \mathcal{L}_{\text{DDPM}}=\mathbb{E}_{z\sim\mathcal{E}(x),y,\epsilon\sim\mathcal{N}(0,1)}[||\epsilon-\epsilon_{\theta}(z_t,t,\tau_{\theta}(y)||_2^2]
% \end{equation}
% SD appends several cross-attention layers to inject the text condition into the latent $z_t$. In the inference stage, the noisy latent $z_T$ is sampled from Gaussian noise, and the final latent $z_0$ produces the image $x$ through $x=\mathcal{D}(z_0)$. 

\textbf{Cross Attention Layers}. SD utilizes the cross-attention mechanism to perform text guidance. For $l$-th cross-attention layer, the intermediate noise latent $z_t$ is projected into queries by a linear layer $f^l_Q(\cdot)$, while text embeddings $\mathcal{E}_{text}(y)$ from CLIP text encoder are projected into keys and values by linear layers $f^l_K(\cdot)$ and $f^l_V(\cdot)$, respectively. The cross-attention maps $\mathcal{A}$ refer to the product between queries and keys:
\begin{equation}
    \mathcal{A}^{l,t}=\text{Softmax} (Q^{l,t}(K^{l})^T/\sqrt{d})
\end{equation}
\begin{equation}
    Q^{l,t}=W^{l}_Q f^{l}_Q(z_t),~K^{l}=W^{l}_Kf^{l}_K(\mathcal{E}_{text}(y))
\end{equation}
where $W^l_Q,W^l_K$ are learnable matrices, $\sqrt{d}$ is the dimension of the feature to normalize softmax values.
%\textbf{Cross Attention Controls.} Prompt-to-Prompt \cite{hertz2022prompt} proposed to manipulate the spatial layouts of images by controlling the cross-attention map.

\subsection{Evaluation Metrics Discussion}
\begin{figure}[tb]
\centering
\includegraphics[width=0.6\columnwidth]{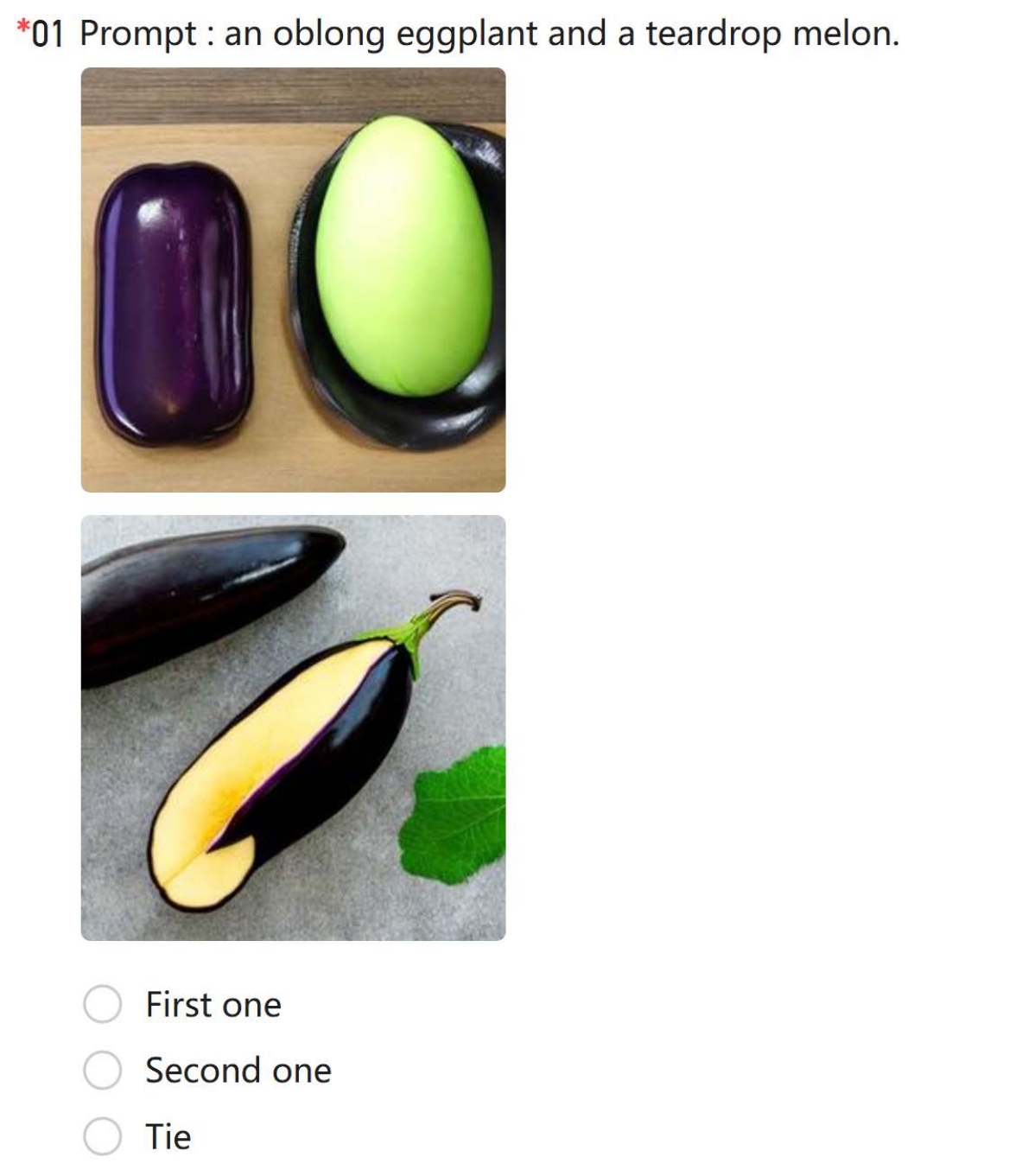}
\caption{A screenshot of the human evaluation of text-image alignment. Evaluators are presented with two images by PiCo and another method of baselines to choose which one better reflects the target prompt or ``tie".}
\label{fig:alignment}
\end{figure}

The unreliability of common testing metrics in attribute binding tasks had been discussed in previous study \cite{zhuang2024magnet}. Thus we mainly rely on human evaluation as our quantitative validation. In Tab. \ref{tab:auto}, we also report two widely used metrics, CLIP score \cite{hessel2021clipscore} and VQAScore \cite{li2024genai}. These scores are obtained on 400 prompts from T2I-Compbench dataset. Our PiCo achieves the highest VQAScore which shows a better text-image alignment. In terms of CLIPScore, we observe all methods obtain similar scores, which also indicate the unreliability of this metric. Regarding the human evaluation, we give a screenshot of the human evaluation in terms of text-image alignment Fig. \ref{fig:alignment}.

\subsection{More Implementation Details}
\textbf{Baselines Setup}. In this section, we provide additional experimental setups for baselines. We assign the same parser, Stanza\cite{qi2020stanza}, for both Attend-and-Excite\cite{chefer2023attendandexciteattentionbasedsemanticguidance} and BoxDiff \cite{xie2023boxdiff}, to extract target token indices from text prompts automatically. We ask GPT-4o \cite{OpenAI-GPT-4o} to generate the bounding boxes of each image as the additional condition requirement of BoxDiff. We use the following prompts to prompt GPT-4o:

\textit{I have a text prompt for text-to-image generation, which contains objects that I have provided. I need you to assign reasonable positions to these objects on a 512x512 image and represent them with the coordinates of the bottom left and top right corners of the bounding box. For example, I give ``a black cat and a white dog", the objects are ``cat" and ``dog", and the bounding boxes are [50, 50, 220, 250], [260, 100, 460, 460].}

\begin{table}[tb]
\caption{Automatic comparison on two widely used metrics. Note that these results are not reliable for text-image alignment.}
\centering
\rowcolors{4}{gray!10}{gray!10}
\begin{tabular}{ccc}
\toprule
     \multirow{2}{*}{Model}&\multicolumn{2}{c}{Benchmarks}\\
    \cmidrule(r){2-3}&VQAScore $(\uparrow)$&CLIPScore $(\uparrow)$\\
    \hline
    Stable Diffusion&0.45&16.24\\
    Attend-and-Excite&0.47&16.35\\
    BoxDiff&0.42&16.22\\
    GORS&0.45&16.32\\
    Magnet&0.43&16.29\\
    \rowcolor{gray!50}
    Ours&\textbf{0.50}&16.26\\
    \rowcolor{white}
\bottomrule
\label{tab:auto}
\end{tabular}
\end{table}

GORS \cite{huang2023t2i} provides fine-tuned LoRA weights for different types of prompts (color, texture, shape, and complex). For a fair comparison, we employ a specific LoRA weight of GORS according to the category of the current prompt in both quantitative and qualitative results. We use the default setting of Stable Diffusion \cite{rombach2021highresolution} and Magnet\cite{zhuang2024magnet}.

% The prompt we gave to GPT-4o to generate long prompts is as follows:

% \textit{Please help me write some text prompts for text-to-image generation. Each prompt should contain at least three objects, with each object described by an adjective (preferably a color). Additionally, there should be some positional relationships between the entities, for example: ``A yellow bus next to a green tree and a red mailbox”, ``An orange cat on a gray roof overlooking a white fence”. Now, please give me 15 such prompts.}

\subsection{Additional Ablation Study}
\begin{figure}[tb]
\centering
\includegraphics[width=\linewidth]{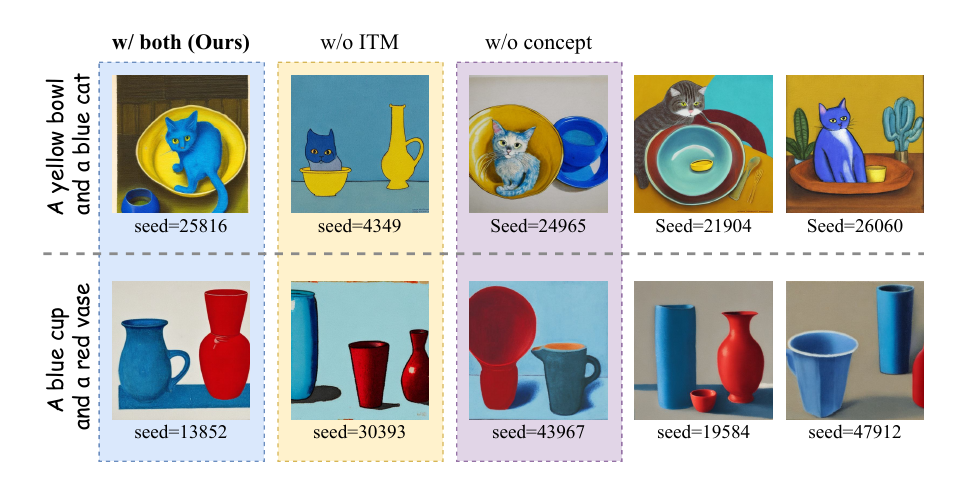}
\caption{Ablation study on two types of score. Given a candidate set with 5 unique noises, column 2 is the selected noise by concept scores, column 3 is chosen by the ITM score. Using both types of scores is able to select the most suitable noise in column 1.}
\label{fig:seed_score}
\end{figure}

% 去掉concept score，只保留ITM socre/反之
\textbf{Noise Score}. In the main paper, we assess the quality of each noise as the summed score of the ITM score and concept scores. In this section, we discuss the choice of using each score independently to measure the noise quality. In Fig. \ref{fig:seed_score}, columns 1-3 are the selected noises via both scores (column 1), only the concept score (column 2), and only the ITM score (column 3). Relying solely on the ITM score is inaccurate in measuring the existence of fine-grained visual concepts, while using only the concept score leads to the problem of quality degradation (e.g., split objects). The combination of two kinds of scores is reliable for the selection of initial noise for better global quality and local information to encourage each concept.

\begin{figure}[tb]
\centering
\includegraphics[width=\columnwidth]{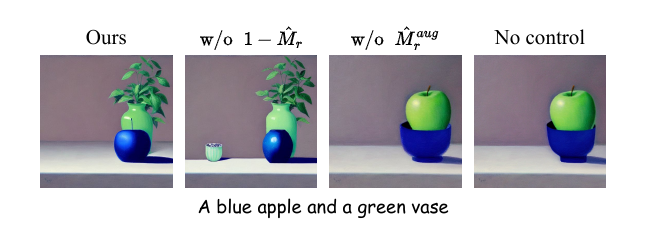}
\caption{Ablation study on two types of masks. Without the exclusive mask, two objects stick together and cannot be distinguished well. Without the concept mask, the controlled result resembles the original image. It is necessary to use two masks to improve the synthesis quality.}
\label{fig:inter-intra}
\end{figure}

\begin{figure}[bt]
\centering
\includegraphics[width=0.9\columnwidth]{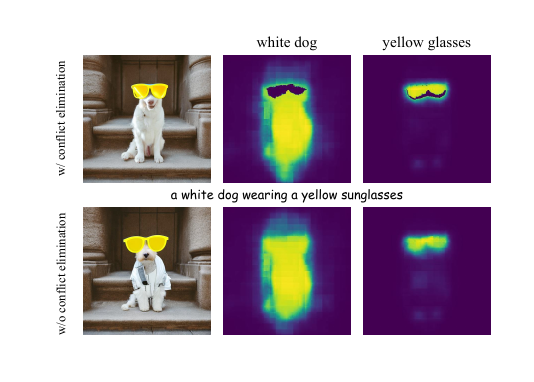}
\caption{Ablation study on the conflict elimination. We show the images and masks generated by our referring mask control module. The pixel in overlapped areas can be assigned to the concept with higher activations to avoid conflict. The object ``dog" in the top row with conflict elimination is realistic while the bottom row shows cloth-like artifacts.}
\label{fig:conf}
\end{figure}

\begin{figure*}[bt]
\centering
\includegraphics[width=\textwidth]{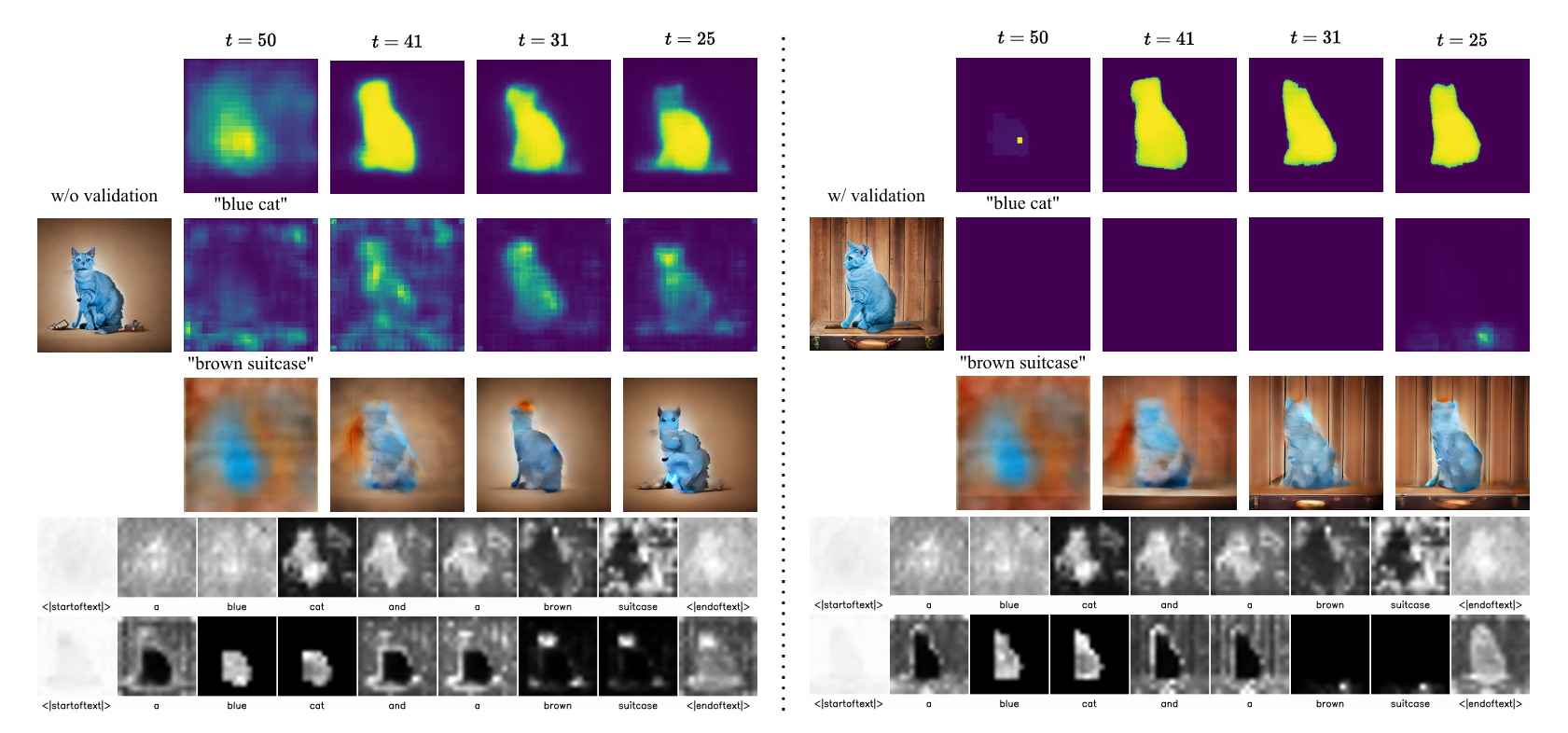}
\caption{Ablation study of the mask validation. (Left) Without the validation, sparse activations on the segmentations of the object ``brown suitcase" entangles the visual representation of another concept, resulting in strong artifacts in the intermediate images. (Right) We apply the mask validation scheme to prevent the cross-attention interaction of the denoising U-Net from being affected by unreliable masks with noisy activations.}
\vspace{-2mm}
\label{fig:validation}
\end{figure*}

\textbf{Concept Mask and Exclusive Mask}. In Fig. \ref{fig:inter-intra}, we investigated the effectiveness of exclusive mask ($1-\hat{M}_r$) and concept mask ($\hat{M}^{aug}_r$) in the referring mask control module. If no mask is used at all, it degrades to the original SD that lacks attribute binding ability (the last column). In column 3, the sole use of the exclusive mask shows limited improvement in the generated image, indicating the importance of the concept mask in guiding the cross-attention interaction. In column 2, the use of the concept mask significantly improves the binding, presenting desired concepts. However, the image can be observed with a small green cup on the left side, which can be caused by unexpected noise activations on the cross-attention maps. With both concept and exclusive masks, the control is effective and reliable for proper attribute binding and removing irrelevant objects.
% \textbf{Concept Mask and Exclusive Mask.} In Fig. \ref{fig:inter-intra} we investigated the effectiveness of exclusive mask ($1-\hat{M}_r$) and concept mask ($\hat{M}^{aug}_r$) for single visual concept. If these two masks are not used at all, it will degrade to the original SD, creating some poor attribute binding ability (see last column). Similarly, if $\hat{M}^{aug}_r$ is not used, it can be observed that the control is not strong enough to alleviate such phenomenon. Without the exclusive mask, it can be observed that a small green cup appears on the left side of the image, which indicates that there are some wrong activation areas on the cross-attention maps. With both the concept mask and exclusive mask, the effective control makes proper attribute binding and avoids irrelevant objects.

\textbf{Conflict Elimination}. Fig. \ref{fig:conf} provides an example of conflict elimination. Obviously, two objects ``white dog" and ``yellow sunglasses" have overlapped activations. In the bottom row, the visual features cannot distinguish the ownership of pixels in the conflict area during the cross-attention interaction, resulting in out-of-distributed representations and showing severe artifacts in the generated image. With conflict elimination in the top row, two masks of concepts do not overlap and produce a more reasonable generation result.

% 去掉mask validation
\textbf{Mask Validation}. In the main paper, we propose to discard CLIPSeg segmentations if they have low confidence or sparse activations. As shown in Fig. \ref{fig:validation}, when disabling the mask validation, each segmentation mask was retained, but in the early stage (i.e. $t=50$), the segmentation result of token ``brown suitcase" is not significant and consists of some messy highlights. Reckless use of these random masks can disrupt the semantic information of objects, resulting in the missing objects problem (see left). It can be seen in the right part, that the mask validation can filter out those segmentation results with low confidence. Although the token ``brown suitcase" seems to have never been effective as it keeps not activated for the whole process, the semantic information possessed by the original region is sufficient to generate the described object without interference from the other token.  

\textbf{Hyperparameter $r_s$ for Candidate Set Size}. In order to select suitable noises to generate $N$ images, we introduce a hyperparameter $r_s$ to construct the candidate set. In Tab. \ref{tab:ablation}, we fix the inference step $T_s$ to 5 and study three settings $r_s=5,10,15$ via human evaluation. FID is not reported since the noise selection module does not affect image quality. Compared to no noise selection, the result of $r_s=5$ slightly alleviates the missing problem as the ``None" rate is declined. However, the ``Both" rate is close to the original setting. One possible explanation is that a small set lacks suitable noises to generate multiple visual concepts. This can be validated when increasing the ratio $r_s$ to 10, the "Both" rate jumps from 28.8\% to 36.2\%, and the ``None" rate decreases from 29.1\% to 13.4\%, which shows the effectiveness of our noise selection under a relatively large set size. However, blindly expanding the candidate set size by setting $r_s=15$ will not increase the vote in ``Both" accordingly. This comparison also demonstrates the robustness of assessing the noise quality. Due to equipment limitations, we assess 5 different noises in parallel, while we emphasize that the runtime can be further reduced by using a larger number of parallels. To conclude, we set $r_s=10$ to balance time and performance. 

% \textbf{Seed Set $Q$.} We conducted qualitative ablation experiments on the size $Q$ of the seed set. We only used the seed selection module and disabled the referring mask control module. We did not report FID in the results of $Q$ because the seed selection module does not affect image quality. We set $Q=5/10/15$ separately and the inference step $T_s$ is fixed to 5. In Tab. \ref{tab:ablation}, setting $Q$ to 5 did not significantly improve the reporting rate of ``Both", but successfully reduced the reporting rate of ``None". This may caused by the difficulty of seeds suitable for generating two single visual concepts with a small set. When increasing $Q$ from 5 to 10, the rate of "Both" increased from 28.8\% to 36.2\%, while ``None" decreased from 29.1\% to 13.4\%, which proves the effectiveness of our seed selection. As increasing $Q$ to 15, the rate of ``Both" has slightly increased, but there is a significant increase in the runtime of the seed selection module which increases by nearly 52.6\%. It should be noted that due to GPU memory limitations, we can only calculate up to 5 seeds simultaneously in the experiment. Using a larger number of parallelism can significantly reduce its runtime. We set $Q$ to 10 by balancing time and performance.

\begin{figure*}[bt]
\centering
\includegraphics[width=0.9\textwidth]{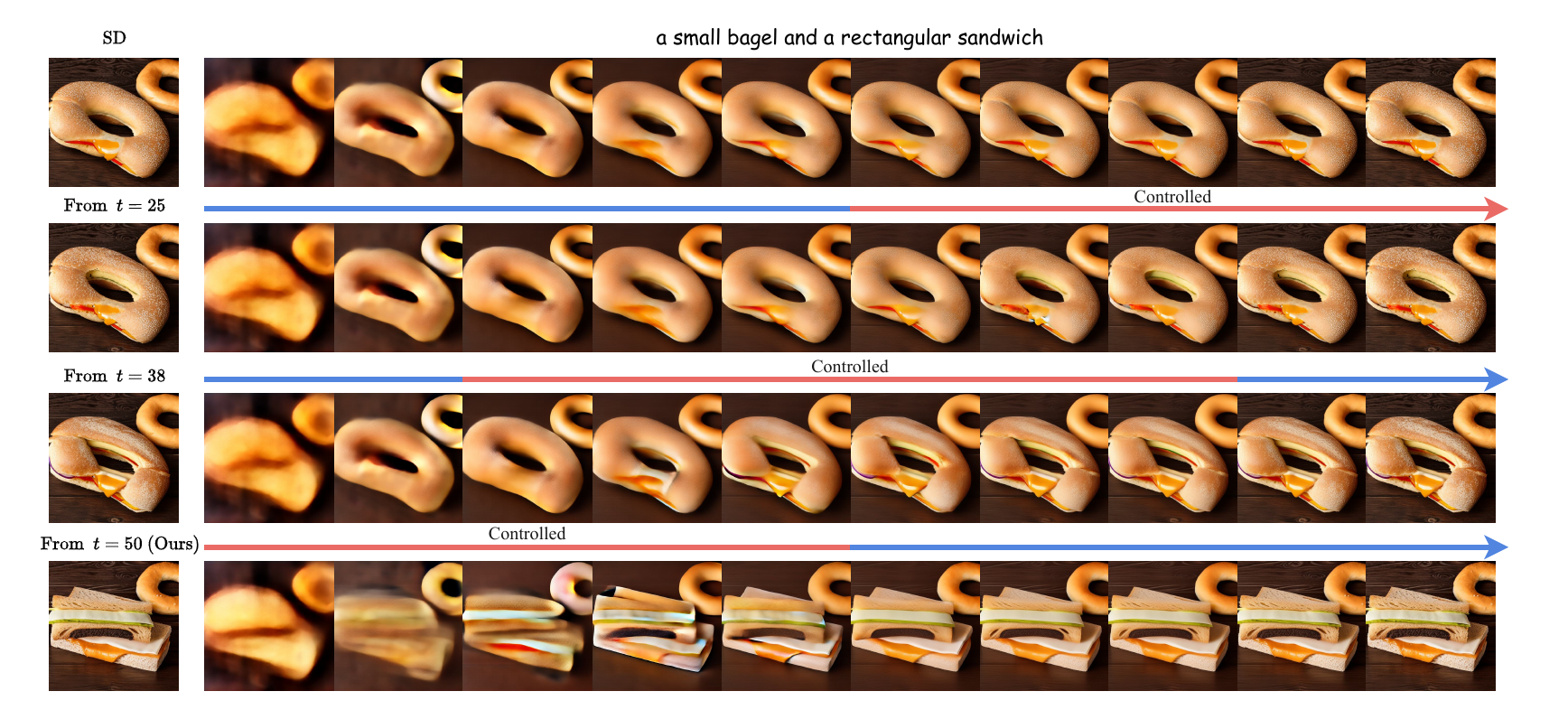}
\caption{Ablation study of different denoising stages to apply the referring mask control. The intervene in the latter (row 2) and the middle (row 3) stages of denoising change fine-grained visual contents in local areas, such as the sandwich's sauce and the texture of crust. To modulate the global information of the concept, we deploy the mask control at the early denoising stage (row 4), which improves the image structure, especially for the shape attribute.}
\label{fig:control_pos1}
\end{figure*}

\begin{figure*}[bt]
\centering
\includegraphics[width=0.9\textwidth]{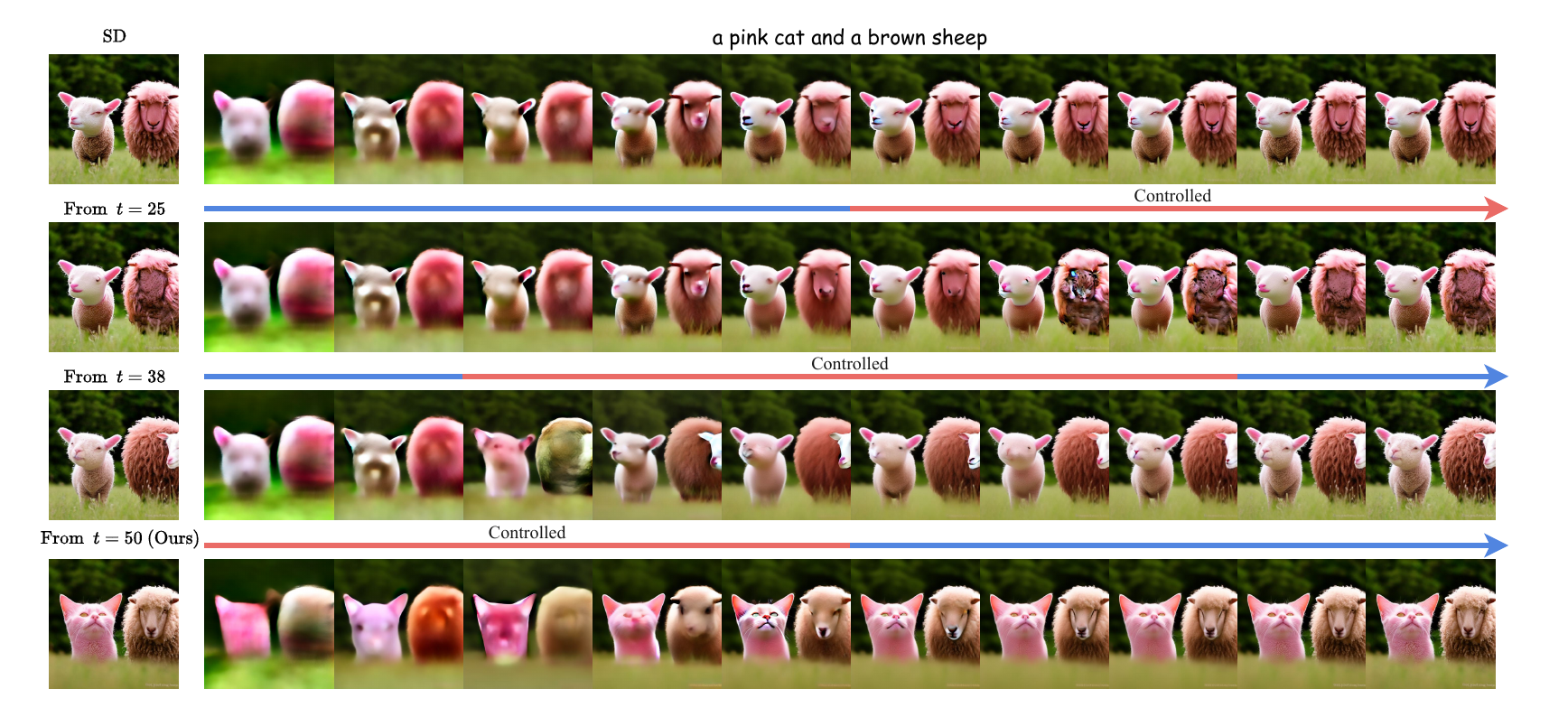}
\caption{Another example of the ablation study on the referring mask control stage.}
\label{fig:control_pos2}
\end{figure*}

% 只在后25步做干预，因为高维细粒度信息应该在后期生成（不过效果应该是不好的）
\textbf{Mask Control Stage}. In the main paper, we deploy the referring mask control in the early denoising stage, where the image layout is most determined. In Fig. \ref{fig:control_pos1} and Fig. \ref{fig:control_pos2}, we explore the effects of different denoising spans to apply the control: the early stage $t\in[50,25]$, the middle stage $t\in[38,13]$, and the latter stage $t\in[25,0]$. In row 2 of the first example, we apply our mask control at the later stage from $t=25$ to $0$. This design is ineffective and results in only subtle changes such as the sandwich's sauce and crust. This also verifies that the latter denoising stage affects the fine-grained information of the image. The second example even presents catastrophic visual representations on the intermediate images in row 2, as the sheep's face is destroyed. A similar discovery is observed in row 3 of two examples when controlling the cross-attention maps from $t=38$ to $13$. Neither the mask control is applied in the middle nor the latter stages, the generated images resemble SD's results. Conversely, early intervention ensures a significant difference in image layout. The results in row 4 (Ours) are visually appealing and faithful to the given prompts, even unconventional concepts like ``rectangular sandwich" and ``pink cat".

\begin{table*}[tb]
\caption{Ablation studies on several crucial hyperparameters in the noise selection and the referring mask control modules.}
\centering
\begin{tabular}{cccccc}
\toprule
     \multirow{2}{*}{Module}&\multirow{2}{*}{Parameter}&\multicolumn{3}{c}{Human Annotation}& \multirow{2}{*}{FID $(\downarrow)$}\\
    \cmidrule(r){3-5}& &Both $(\uparrow)$&Single&None $(\downarrow)$&\\
    \hline
    \rowcolor{gray!10}
    &No Seed Selection&27.8&38.7&33.5&-\\
    \rowcolor{gray!10}
    \multirow{-2}{*}{Seed Selection}&Seed Set $r_s$= 5/\textbf{10}/15&28.8/\textbf{36.2}/37.5 &42.1/\textbf{45.4}/46.0&29.1/\textbf{18.4}/16.5&-\\
    \hline
    \rowcolor{gray!10}
    &$T_c$ = 0 (No control)&28.4&38.6&33.0&41.65\\
    \rowcolor{gray!10}
    &$T_c$ = 5/\textbf{25}/50&29.2/\textbf{37.8}/15.0&39.5/\textbf{42.5}/33.2&31.3/\textbf{19.7}/51.8&38.25/\textbf{35.74}/46.14\\
    \rowcolor{gray!10}
    \multirow{-3}{*}{Mask Control}&$\gamma$=1/\textbf{15}/50 &30.4/\textbf{37.8}/10.9&40.8/\textbf{42.5}/32.9&28.8/\textbf{19.7}/56.2&40.08/\textbf{35.74}/48.33\\
    \rowcolor{white}
\bottomrule
\label{tab:ablation}
\end{tabular}
\end{table*}

\begin{table}[tb]
\small
\caption{Ablation study on the percentile and the hyperparameter $\delta$ in the noise selection module.}
\centering
\begin{tabular}{ccccc}
\toprule
    &Value&Both/Single$\uparrow$&None$\downarrow$&DINO$\uparrow$\\
    \hline
    \rowcolor{gray!10}
    &50&76.7&24.3&0.547\\
    % \rowcolor{gray!10}
    % &70&78.0&22.0&0.567\\
    \rowcolor{gray!10}
    &100&75.9&25.1&0.542\\
    \rowcolor{gray!50}
    \multirow{-3}{*}{percentile}&\textbf{90 (Ours)}&\textbf{80.2}&\textbf{19.8}&\textbf{0.659}\\
    \hline
    \rowcolor{gray!10}
    &1&80.5&21.3&0.635\\
    \rowcolor{gray!10}
    &10&67.5&32.5&0.503\\
    \rowcolor{gray!50}
    \multirow{-3}{*}{$\delta$}&\textbf{2 (Ours)}&\textbf{80.2}&\textbf{19.8}&\textbf{0.659}\\
\bottomrule
\vspace{-4mm}
\label{tab:lambda_sigma}
\end{tabular}
\end{table}

\textbf{Control Step $T_c$ and Augmentation Strength $\gamma$}. In Tab. \ref{tab:ablation}, we study two crucial hyperparameters of the referring mask control module. We also assess FID-score on 5k image-text pairs from MS-COCO \cite{lin2014microsoft} val set. The first hyperparameter $T_c$ determines how many timesteps to manipulate the cross-attention maps. When setting $T_c=5$, the human annotated results indicate a slight improvement against no control. However, the full control of $T_c=50$ can be excessive and lead to the lowest rate of ``Both". Its FID score even gets worse, showing the collapse of image quality. Empirically, we set $T_c=25$ to achieve the best performance on both metrics of text alignment and image quality. The second hyperparameter $\gamma$ controls the augmentation strength of the concept mask. According to the quantitative study, the small value of $\gamma=1$ is effective, but not much as the result of $\gamma=15$. On the other hand, the excessive strength $\gamma=50$ affects the performance of referring mask control. We set $\gamma=15$ as an appropriate value to augment the concept mask.

\textbf{Seed Selection Hyperparameter}. In Tab. \ref{tab:lambda_sigma} we conduct ablation study on extra hyperparameters include the percentile and $\sigma$ used in the noise selection module. Using the $50$-th percentile, the highlight area of image can not be captured, thus causing performance degradation. As for the $100$-th percentile, it is equal to get the maximum value but it cannot represent the significance of the existence of an object as the rate of ``Both/Single" show a lower value than setting to $90$ (Ours). As for $\sigma$, setting $\sigma$ to 1 is equivalent to treating the two parts of formula 4 equally without assigning weights as it will cause a higher rate of missing objects. While setting $\sigma=10$, the concept score will be entirely determined by the first term of formula 4, resulting in significant performance degradation. We set $\sigma=2$ to emphasize the role of the maximum value without losing the influence of the average value.

% \textbf{Control Step $T_c$ and Augmentation Ratio $\gamma$.} In the main text, we have already demonstrated the effects brought by different $T_c$, and here we further provide more ablation experiments. Compared with the ablation experiment in the main text, the results of the human evaluation also confirm our conclusion. Setting $T_c$ to 5 did not bring any improvement in effect, while when $T_c$ was set to 25, the proportion of ``Both" increased from 28.4\% to 37.8\%, while the proportion of ``None" decreased from 33.0\% to 19.7\%; Similar to the situation in the main text, setting $T_c$ to 50 leads to a collapse in image quality, with the proportion of ``None" skyrocketing to 51.8\%. At the same time, we tested FID-5K on the COCO2017\cite{lin2014microsoft} val set to provide more objective support for the experimental conclusions. We also showed the runtime of $T_c$, which proved that mask control would bring a slight time overhead. Also, we report some quantitative results about $\gamma$ in Tab. \ref{tab:ablation}, and these data also support our conclusion in the main text. We didn't report runtime of $\gamma$ cause it has no impact on runtime. 

\begin{figure}[tb]
\centering
\includegraphics[width=0.9\columnwidth]{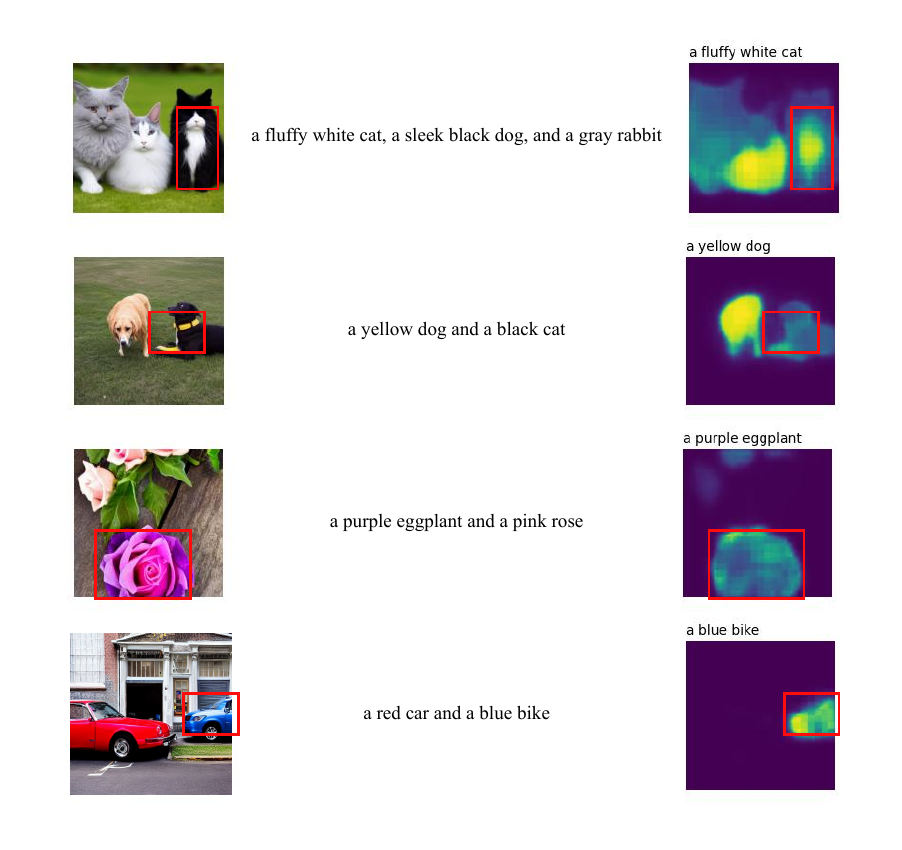}
\caption{The bias of the CLIPSeg model towards color. Prompted by a specific concept, CLIPSeg's segmentations include unexpected entities which only match the color instead of the target object in text referents (marked in the red box). For example, in row 2, CLIPSeg labels the yellow dog leash when asked to segment ``yellow dog".}
\label{fig:clipseg_color}
\end{figure}
\subsection{Limitation}
Although the proposed PiCo is an effective method to address the attribute binding issue, it must be acknowledged that our approach has some limitations.

\textbf{Dependence on the Performance of the Parser}. In both modules, we need to extract visual concepts from the input text prompts, such as ``a pink dog and a blue mouse" to obtain the first concept ``pink dog" and the second concept ``blue mouse". We have adopted an off-the-shelf dependency parsing module in Stanza Library and construct syntax trees using NLTK and found it works well for most simple text prompts. However, errors are prone to occur when faced with complex text prompts. For example, given ``Rice with red sauce with eggs over the top and orange slices on the side", the obtained concept set are ``red sauce", ``eggs", ``top", and ``orange slices", that has neglected the expected concept ``Rice" but incorrectly extracted an invalid noun ``top". 
% In fact, in our experiment, the more complex the text prompts, the worse the parsing results. Inaccurate parsing results not only affect our PiCo but also other methods based on text prompt analysis such as Attend-and-Excite, BoxDiff, Magnet, etc. Although many methods solve the above problems by using user manual specification, it is very tedious and time-consuming when dealing with large amounts of data and complex text.

\textbf{Attribute Bias of CLIPSeg}. In the experiment, we discover an interesting phenomenon that emerges in the CLIPSeg model: a bias towards specific color. As illustrated in Fig. \ref{fig:clipseg_color}, despite the objects marked in red boxes not being the entities with the corresponding color attributes in the prompts, CLIPSeg tends to include the areas associated with these objects in the segmentation results for entities, such as ``a yellow dog" and ``a blue bike." In the former case, the object is not a ``dog," and in the latter, it is clearly not a ``bike," yet both maintain consistency in terms of the attribute (color). This bias in CLIPSeg can be interpreted as a ``compromise" approach. Although it may appear logically reasonable, it is detrimental to our objectives. A potential solution is calculating IoU to delineate color regions and entity regions, thereby filtering out areas where color and entity attributes are inconsistent.

\textbf{Hyperparameters}. There is a substantial number of hyperparameters in our algorithm, including several thresholds and the augmentation factor. Although the experimentally set values perform well in most cases, rigidly setting these hyperparameters lacks adaptability to different situations. Supporting self-adaptive hyperparameters is one of the future directions for improvement.

\begin{figure}[bt]
\centering
\includegraphics[width=0.8\columnwidth]{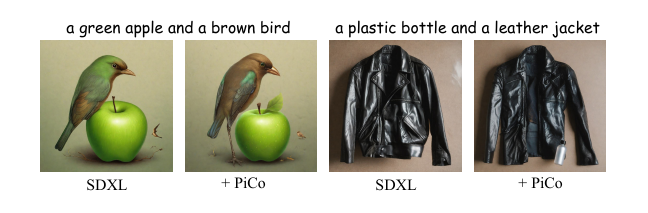}
\caption{PiCo is compatible with other base models.}
\label{fig:sdxl}
\end{figure}

\subsection{Extensions}In Fig. \ref{fig:sdxl} we combine our PiCo with a well-known advanced model, SDXL. We show that PiCo is compatible to other to-date T2I models. 

\subsection{Additional Qualitative Results}

\begin{figure*}[tb]
\centering
\includegraphics[width=\textwidth]{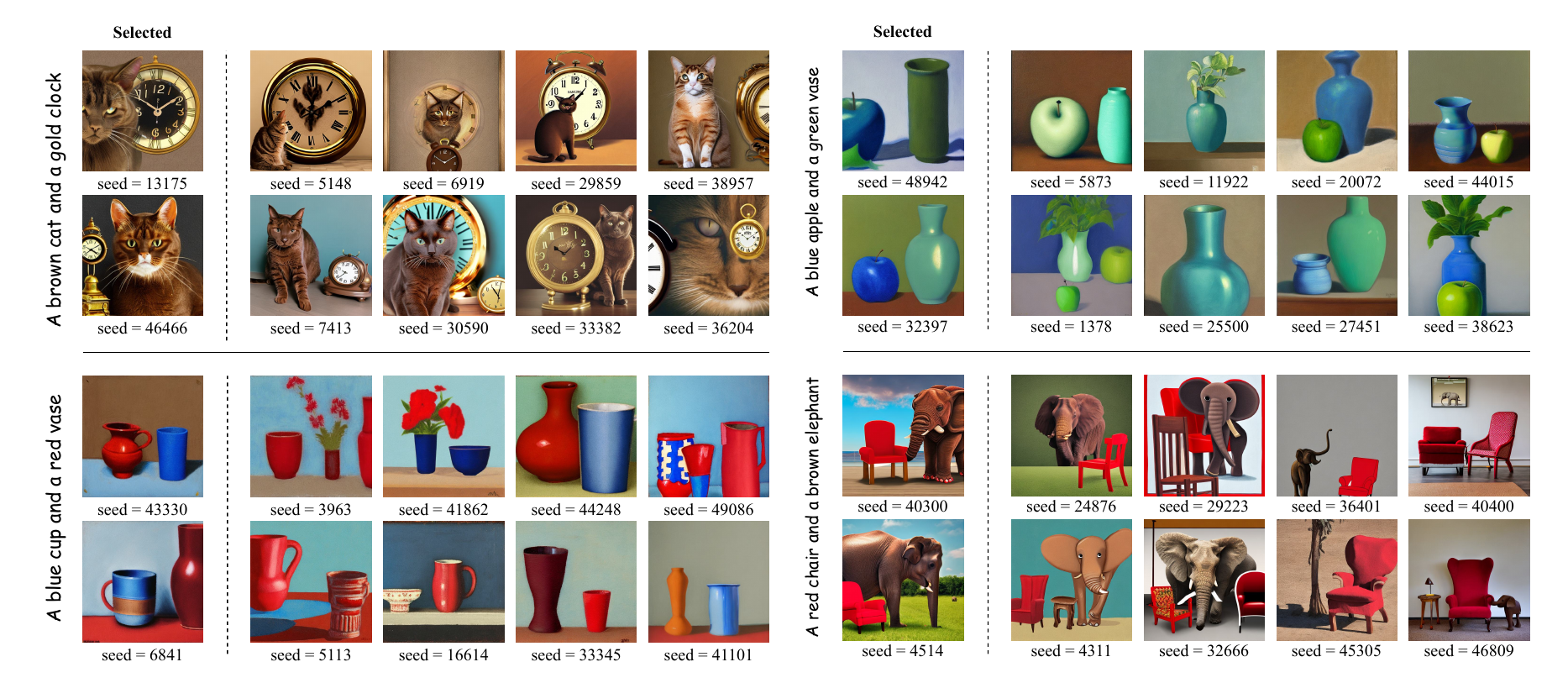}
\caption{Examples of the proposed noise selection module to choose the best 2 noises for the current text prompt from 10 unique noises.}
\label{fig:seed}
\end{figure*}

Fig. \ref{fig:seed} is additional noise selection examples to obtain the best two noises from a fixed size of candidate set.

\begin{figure}[bt]
\centering
\includegraphics[width=0.8\columnwidth]{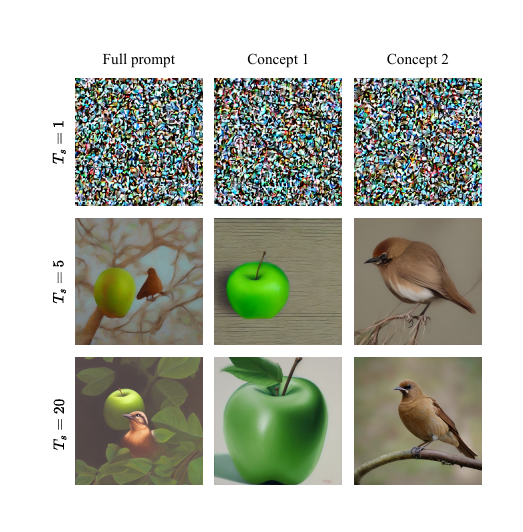}
\caption{Visualization results of different $T_s$, we empirically set $T_s=5$ to balance time and performance. All the images were generated by the text prompt ``A green apple and a brown bird".}
\label{fig:ts}
\end{figure}

In Fig. \ref{fig:ts}, we provide the visualization of the coarse images in the noise selection module with different $T_s$.

\begin{figure*}[bt]
\centering
\includegraphics[width=\textwidth]{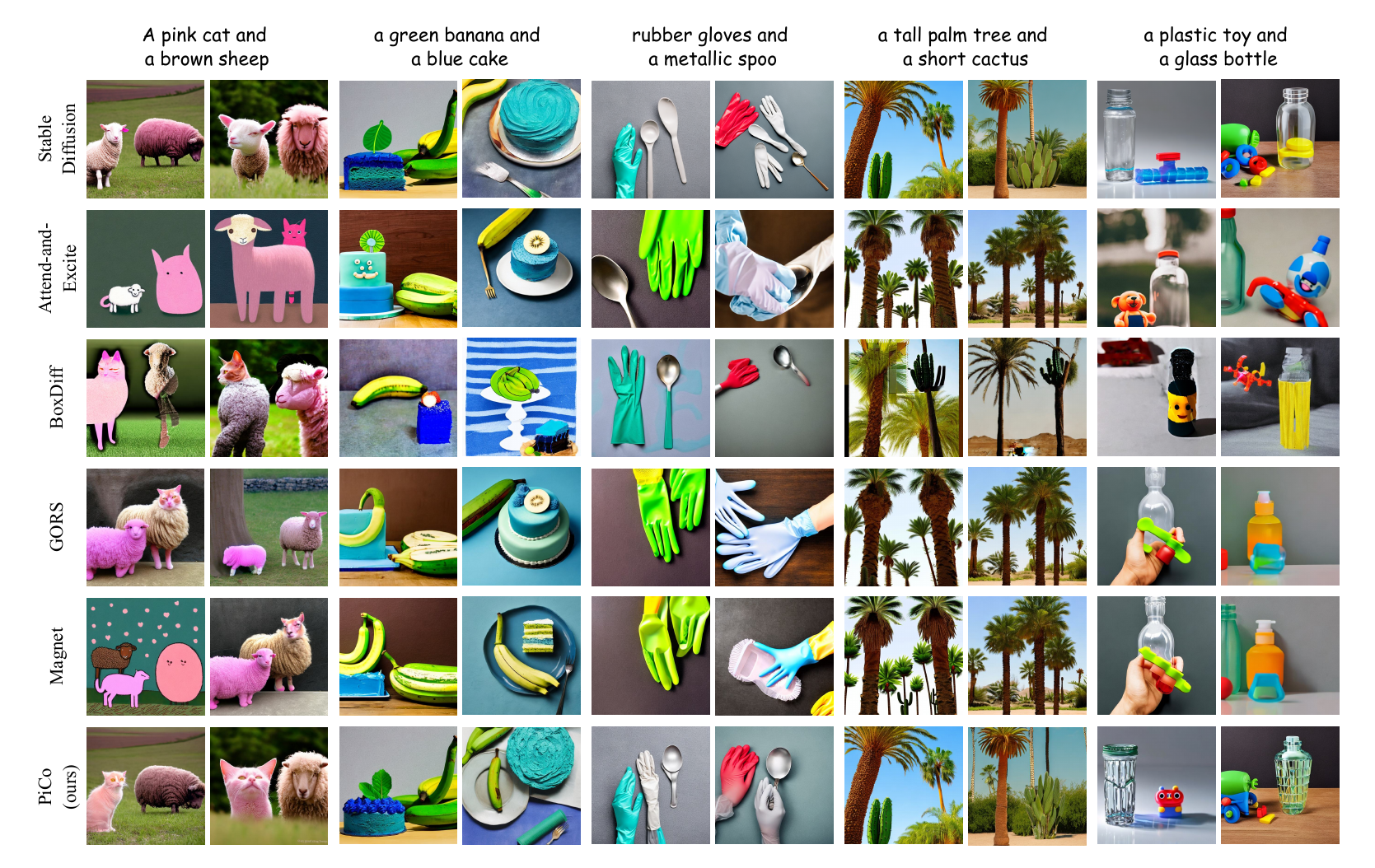}
\caption{More qualitative results on T2I-Compbench dataset for attributes of color, texture and shape.}
\label{fig:qualitative_appendix}
\end{figure*}

\begin{figure*}[bt]
\centering
\includegraphics[width=\textwidth]{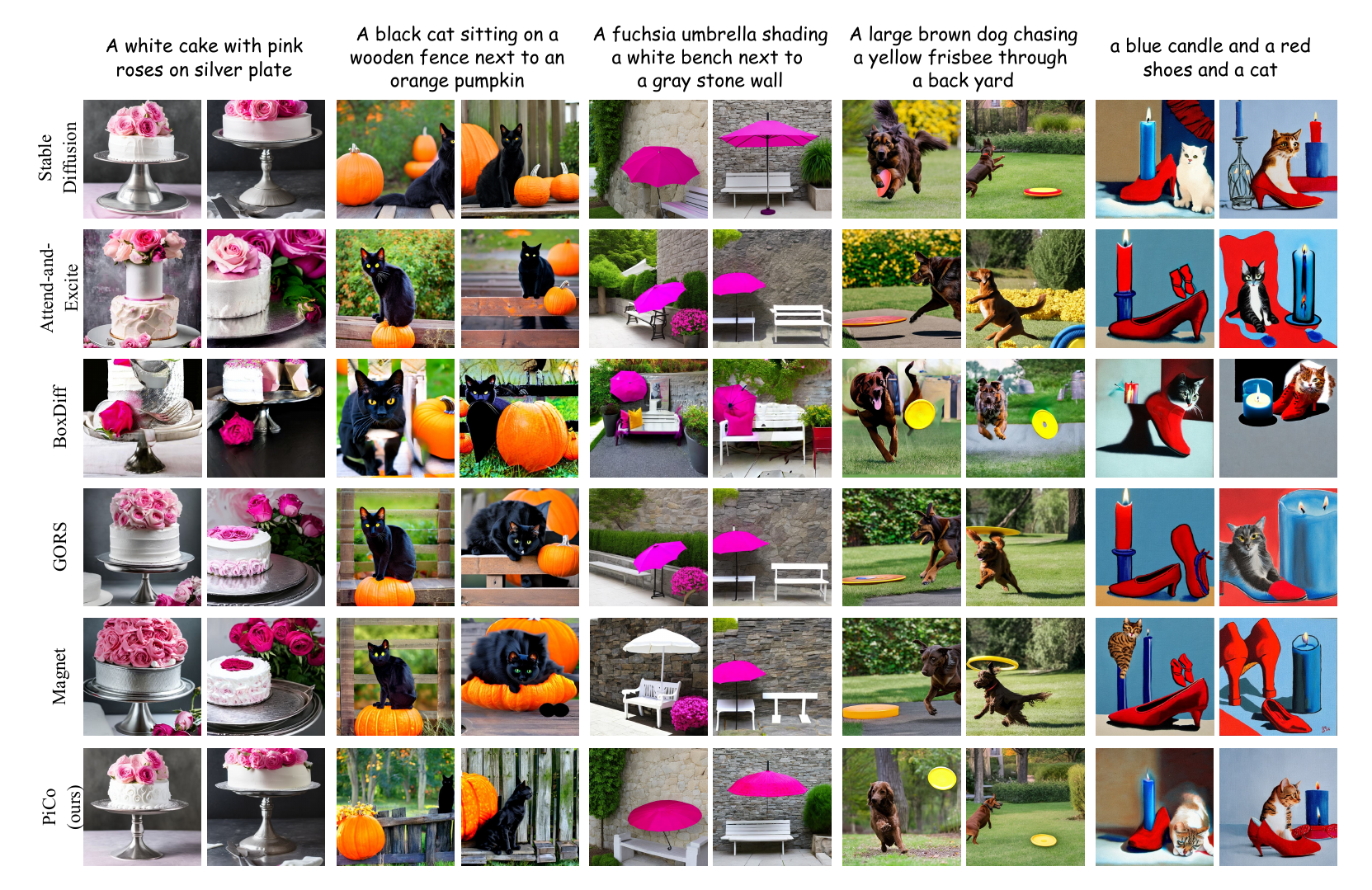}
\caption{More qualitative results on complex prompts. We utilize GPT-4o to generate these long text prompt automatically.}
\label{fig:long}
\end{figure*}

Fig. \ref{fig:qualitative_appendix} and Fig. \ref{fig:long} provide more qualitative comparisons on prompts from the T2I-Compbench dataset and other complex prompts.

\begin{figure*}[bt]
\centering
\includegraphics[width=\textwidth]{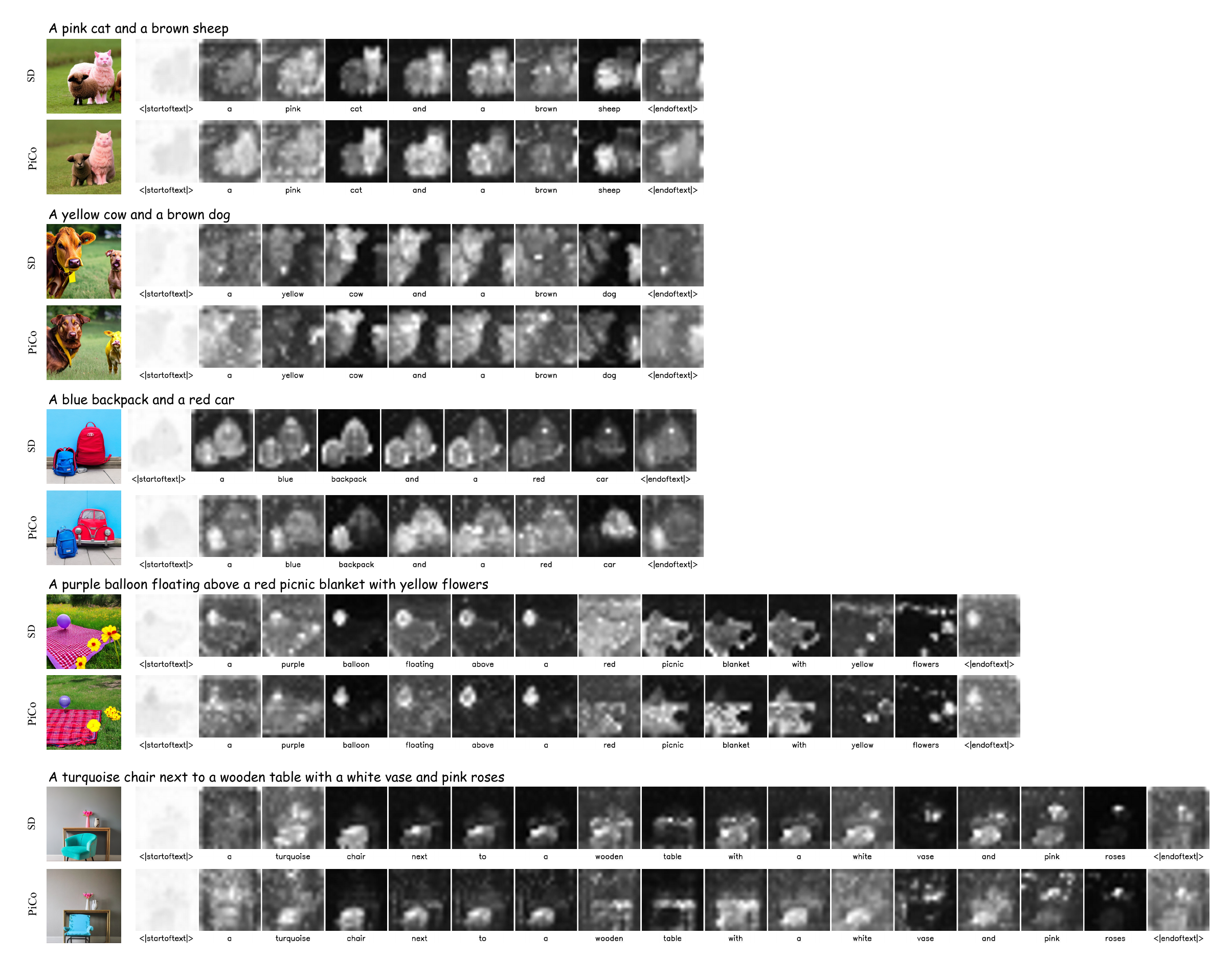}
\caption{Visualization of cross-attention maps to verify the effectiveness of the referring mask control. Compared to the original SD, our proposed PiCo has more aggregated activations (``cat" token in the first example) and disentangled concept representations (``backpack" and ``car" tokens in the third example).}
\label{fig:attn}
\end{figure*}

In Fig. \ref{fig:attn}, we compare the cross-attention visualization between SD and PiCo.

\end{document}